%% file: main.tex
\documentclass[a4paper,11pt]{article}

\input{styles_and_packages/preamble}

\begin{document}

\title{Maybe you don't need a U-Net: convolutional feature upsampling for materials micrograph segmentation}

\author[1, 3]{{Ronan Docherty}
\thanks{ronan.docherty18@imperial.ac.uk}\ \ }

\author[2, 3]{{Antonis Vamvakeros}}

\author[3]{Samuel J. Cooper}

\affil[1]{{\textit{\footnotesize Department of Materials, Imperial College London, London SW7 2DB}}}

\affil[2]{{\textit{\footnotesize Finden ltd, Building R71, Rutherford Appleton Laboratory, Harwell Science and Innovation Campus, OX11 0QX, UK}}}

\affil[3]{{\textit{\footnotesize Dyson School of Design Engineering, Imperial College London, London SW7 2DB}}}

\lhead{\scshape Docherty \textit{et al.}}
\chead{\scshape Feature Upsampling \& Micrograph Segmentation}
\rhead{\scshape Preprint}

\maketitle


\begin{abstract}
\begin{center}
\begin{minipage}{0.85\textwidth}
{\small Feature foundation models - usually vision transformers - offer rich semantic descriptors of images, useful for downstream tasks such as (interactive) segmentation and object detection. For computational efficiency these descriptors are often patch-based, and so struggle to represent the fine features often present in micrographs; they also struggle with the large image sizes present in materials and biological image analysis. In this work, we train a convolutional neural network to upsample low-resolution (\textit{i.e,} large patch size) foundation model features with reference to the input image. We apply this upsampler network (without any further training) to efficiently featurise and then segment a variety of microscopy images, including plant cells, a lithium-ion battery cathode and organic crystals. The richness of these upsampled features admits separation of hard to segment phases, like hairline cracks. We demonstrate that interactive segmentation with these deep features produces high-quality segmentations far faster and with far fewer labels than training or finetuning a more traditional convolutional network.  
}

\end{minipage}
\end{center}
\end{abstract}
\vspace{.2cm}
\begin{multicols}{2}

\section{Introduction}
\label{sec:intro}

Segmentation of micrographs is necessary for most kinds of image analysis in materials science, be it phase quantification \cite{LIB_PAHSE_QUANT, CRACK_DETECTION, KINTSUGI, AL_FOR_BATTERY_SEG, BIPHASE_STEEL}, morphological characterization (particle size distribution \cite{PARTICLE_SIZE_DIST_LIB} \cite{MICRONET_CNN}, inter-particle distance \cite{BINDER_DETACHMENT} \textit{etc.}) or transport simulations (\textit{e.g,} tortuosity factor calculation \cite{TAUFACTOR, TAU2}).
However, it is often very challenging to segment these micrographs accurately: they can display imaging artefacts (low signal-to-noise ratio, pore-back\cite{KINTSUGI}, charging and edge effects in electron microscopy; angular undersampling or beam hardening and photon starvation artefacts in computed tomography\cite{XCT_PRIMER}), include complex or ambiguous phases (cracks, polymorphs), and are fundamentally physically distinct from natural images (\textit{i.e,} typical photographs).

The gold standard for scientific image segmentation - the convolutional neural network (CNN) - often requires a large amount of (often densely) labelled training data \cite{U-NET, SPARSE_U-NET}.
This can be expensive and time consuming to collect, even with recent tools \cite{AL_FOR_BATTERY_SEG, SAM, SAMBA}, and the subsequent models can take a long time to train or fail to generalise when applied to new, unseen systems or stoichiometries.
Researchers often have to change their setups, and labelling, creating and training (or fine-tuning) a CNN each time represents a substantial challenge. 
Adaptable pretrained backbones, like Cellpose \cite{CELLPOSE_2} and MicroNet \cite{MICRONET_CNN}, go some way to fixing this, though training a head network or finetuning still requires time and densely labelled training data.
Classical methods like Otsu thresholding \cite{OTSU}, \textit{k}-means clustering \cite{K-MEANS} or watershed segmentation \cite{WATERSHED_TRANSFORM} require no `training data' and little user input, but can be sensitive to noise \cite{OTSU_NOISE}, are limited in their expressive capacity, and often require multiple rounds of manual preprocessing to work well \cite{OTSU_MULTIPLE_STEPS}.

Trainable or interactive segmentation methods like Weka \cite{WEKA_PAPER} or ilastik \cite{ILASTIK} extract classical features like average local intensities, edge magnitudes and textures, and train a classifier (usually a random forest) to map from those features to user-drawn pixel-labels.
They can account for noise, exposure variation and train quickly, with only sparse labels needed.
This makes them appealing in data scarce fields like materials science, where the imaging setup, phase of interest, processing conditions and stoichiometry can change frequently.

Despite their success, interactive segmentation with classical features can struggle with complex phases in microstructures \cite{HR-DV2}.
Various extensions exist, like `autocontext' (retraining on the output classifier probabilities) \cite{AUTO-CONTEXT} or enforcing experimental volume fractions in the loss \cite{EXPERT_SEG_VFS}, but these are still fundamentally limited by their original feature-set.

Feature foundation models like DINO \cite{DINO}, DINOv2 \cite{DINOv2} and I-JEPA \cite{I-JEPA} are models trained to learn general purpose image features over large datasets. They offer rich semantic features, but describe coarse `patches' of an image (as opposed to pixel-level classical features).
Recently work has been done on upsampling these features\cite{STRIDED, FEATUP, LIFT, HR-DV2, LOFTUP}, as well as on integrating these features into interactive segmentation workflows\cite{HR-DV2}, but practical considerations (memory/time scaling, feature quality) has limited success. 

In this work, we introduce an efficient convolutional feature upsampler and leverage it for interactive segmentation of various multi-phase material systems: a lithium-ion cathode, nickel-superalloys and copper ore. 
The efficient design of the upsampler allows it to be used in an application context, in real-time, and on a laptop GPU. We summarise our contribution in Figure \ref{fig:intro_column}.

\section{Background}
\label{sec:theory}

\subsection{Interactive feature-based segmentation}
\label{sec:theory:trainable_seg}
Interactive segmentation, first popularised in biological image analysis \cite{WEKA_PAPER, ILASTIK, NAPARI-APOC}, trains a classifier to map from a features that describe a given pixel to the pixel's class, over a set of user-defined labels. These features are computed over a range of length scales, $\boldsymbol{\sigma}$, often $\boldsymbol{\sigma} = \{\sigma_{\text{min}}, ..., 2^{n-1}\sigma_{\text{min}}\}$ defining the size of a kernel applied to every pixel. 
They are `classical' in that they are not learned, and reflect our understanding of information useful (and physically significant) for phase segmentation: average colour via Gaussian blurs, edge intensity via Sobel or Laplacian filters, texture via a Hessian filter, \textit{etc.} \cite{WEKA_PAPER, ILASTIK}
The choice of features/filters used - the `feature-set` - produces a $(H, W, N)$ `feature-stack' for a $(H, W)$ image.

Users draw labels onto pixels, indicating them as one of $C$ classes, \textit{e.g.} grain/grain boundary, active material/pore/binder or nucleus/cytoplasm/organelles \textit{etc.}
A classifier $f_{\theta}$ is then trained to map from the $N$ dimensional features of labelled pixels to their class, $f_{\theta}: \mathbb{R}^{N} \rightarrow \mathbb{Z}, \mathbb{Z} \in [1, ..., C]$, and applied to the unlabelled pixels to produce a segmentation.
This classifier is usually a random forest, which is capable of expressing nonlinear relationships not typically captured in the classical features\cite{WEKA_PAPER}.

\subsection{Vision transformers (ViTS)}
\label{sec:theory:vits}

\begin{figure}[H]
\centering
    \includegraphics[width=\linewidth]{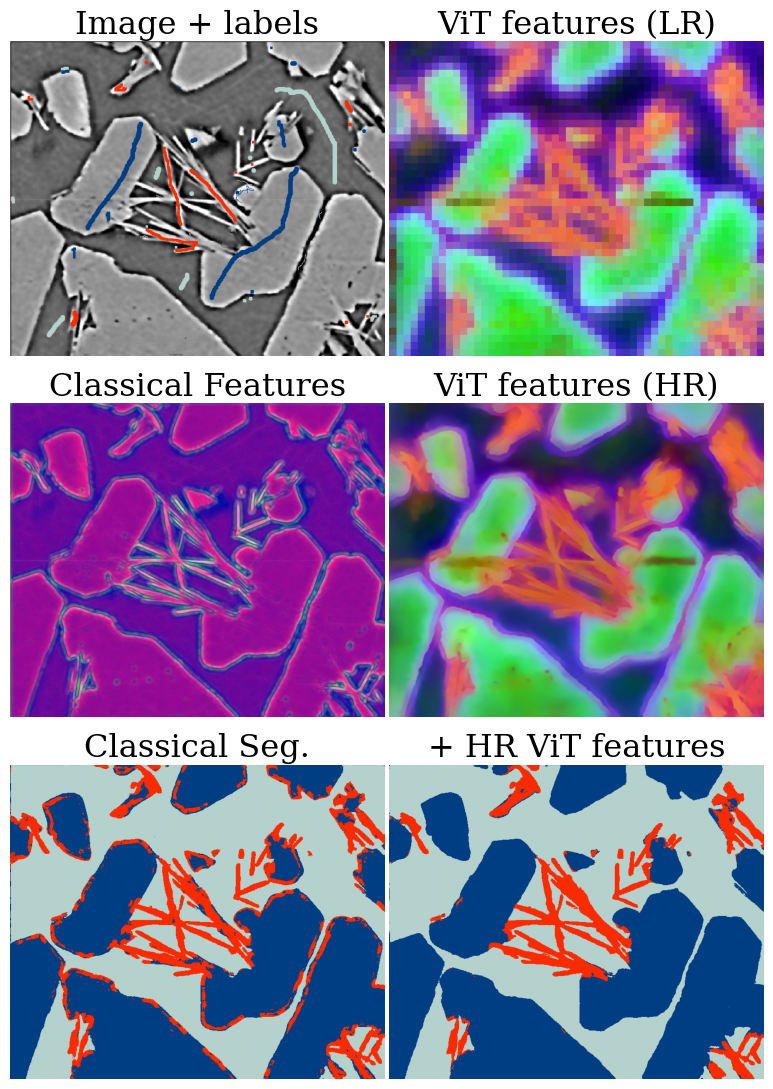}
    \caption{An explanation of our main contributions: we efficiently upsample coarse semantic features from DINOv2\cite{DINOv2} and combine it with pixel level classical features for interactive segmentation. The richness of the ViT features allows the classifier (trained to map from features $\rightarrow$ user labels) to perform more accurate segmentation of the blocky $\alpha-$ and needle-like $\beta-$polymorph phases of an XCT cross-section of glutamic acid \cite{GLUTAMIC_ACID}.}
    \label{fig:intro_column}
\end{figure}

A transformer is a deep neural network that uses `attention', which is the dot-product similarity of learned projections of parts of its input, as its primary mechanism \cite{ATTENTION_IS_ALL_YOU_NEED}.
An input is broken into $T \in \{t_1, t_2, ...\}$ tokens, which are $k$ dimensional vectors and 
the network learns to update the representation of token $t_i$ with respect to all other tokens in $T$ to minimise some overall loss \cite{ATTENTION_IS_ALL_YOU_NEED}.

Transformers were initially applied to sequence modelling tasks, where the input tokens were (approximately) words in a sentence. The all-to-all nature of attention means the computational cost scales $\mathcal{O}(n_t^2)$ where $n_t$ is the number of tokens \cite{ATTENTION_IS_ALL_YOU_NEED}. 
Therefore, to apply transformers to images, inputs had to be broken in non-overlapping patches - usually $16\times16$ or $8\times8\:\text{px}^2$ - as giving each pixel its own token would be computationally impractical \cite{VIT}.

The attention mechanism is fundamentally unordered (\textit{i.e,} invariant under permutations of the input), so additional positional information has to be added to tokens via a positional encoding \cite{ATTENTION_IS_ALL_YOU_NEED}.
This can be simple, like raster order \cite{VIT}, more complex like a sinusoidal encoding \cite{SINUSOIDAL_ENC}, or even learned during training.

Vision transformers have been applied to tasks like classification \cite{VIT}, depth perception \cite{DEPTH_ANYTHING} and segmentation \cite{SAM}, and, due to the favourable scaling of transformers with added data, have quickly reached state-of-the-art results.

\subsection{Feature learning \& DINO}
\label{sec:theory:feature_learning}
Self-supervised learning (SSL) \cite{SSL} is where a network learns representations by mapping between transformations or augmentations of an input, for example learning to fill in missing patches of an image \cite{MAE} - the training target is simply the whole uncovered image.
SSL is attractive because it admits learning without (expensive) human labels, and learns robust representations which can later be used in a supervised context via fine-tuning or adaptor networks \cite{SSL}.

DINO (self-distillation with no labels)\cite{DINO} used a student-teacher framework to learn semantic image features from local-global correspondences in a fully self-supervised framework. They found that this approach learnt (without explicit guidance) semantic decompositions of scenes, and that its features could be used for classification with a linear probe.

DINOv2 \cite{DINOv2} continued this approach at-scale, using more data and a series of techniques (new losses, regularisations, \textit{etc.}) to stabilise the training process.
The model's features were semantically rich and consistent `despite changes of pose, style or even objects'\cite{DINOv2}; they then trained a series of head networks (linear probes or convolutional heads for dense tasks) on top of those frozen features.

\subsection{Feature upsampling}
\label{sec:theory:upsampling}
Features from ViTs are rich, but coarse, and in some domains (like materials science) there may not be enough labelled training data to train a convolutional head network for a given dense task (as in ref. \cite{DINOv2}).
This motivates `feature upsampling': increasing the resolution of these features up to full-pixel resolution or otherwise. These algorithms often use information from the image in order to perform the upsampling. 

Various approaches to ViT feature upsampling have been attempted. One of the first was to reduce the stride of the patch projection layer (\textit{i.e,} moving to overlapping patches): this worked without any further modification or training, but came with increased memory and time cost due to the increased number of tokens, as well as feature blurring \cite{STRIDED, HR-DV2}.

FeatUp \cite{FEATUP} introduced two methods for feature upsampling: a fast, single-pass Joint Bilateral Upsampling filter (JBU) and a slower, per-image implicit approach.
Both were trained by comparing features extracted by a ViT over different views/transformations (flips, zooms, pads) of an input image(s), drawing analogy to Neural Radiance Fields (NeRFs) \cite{NERF}.
The JBU approach is fast, but has steep memory scaling and can suffer from blurring; the implicit approach has sharp features but must be trained separately for each image, which is slow.

LiFT \cite{LIFT} trains a feature upsampling network in a self-supervised fashion, by extracting features of a half-resolution image, upsampling them with a CNN and comparing to the (ground-truth) features of the full-resolution image.
Similar to FeatUp implicit, they use a trained downsampler to extract salient features from the image for upsampling; the approach is lightweight in terms of time and memory. 

LoftUp \cite{LOFTUP} learns a `coordinate-based feature upsampler' (a coordinate transformer) and uses sharp SAM-based \cite{SAM} `pseudo ground truths' for training. This results in sharp upsampled feature maps and fast performance, but with large increases in memory cost as a function of image size, due to every pixel being a token and the quadratic scaling of attention. 

\section{Method}
\label{sec:method}

\begin{figure*}
\centering
    \includegraphics[width=\linewidth]{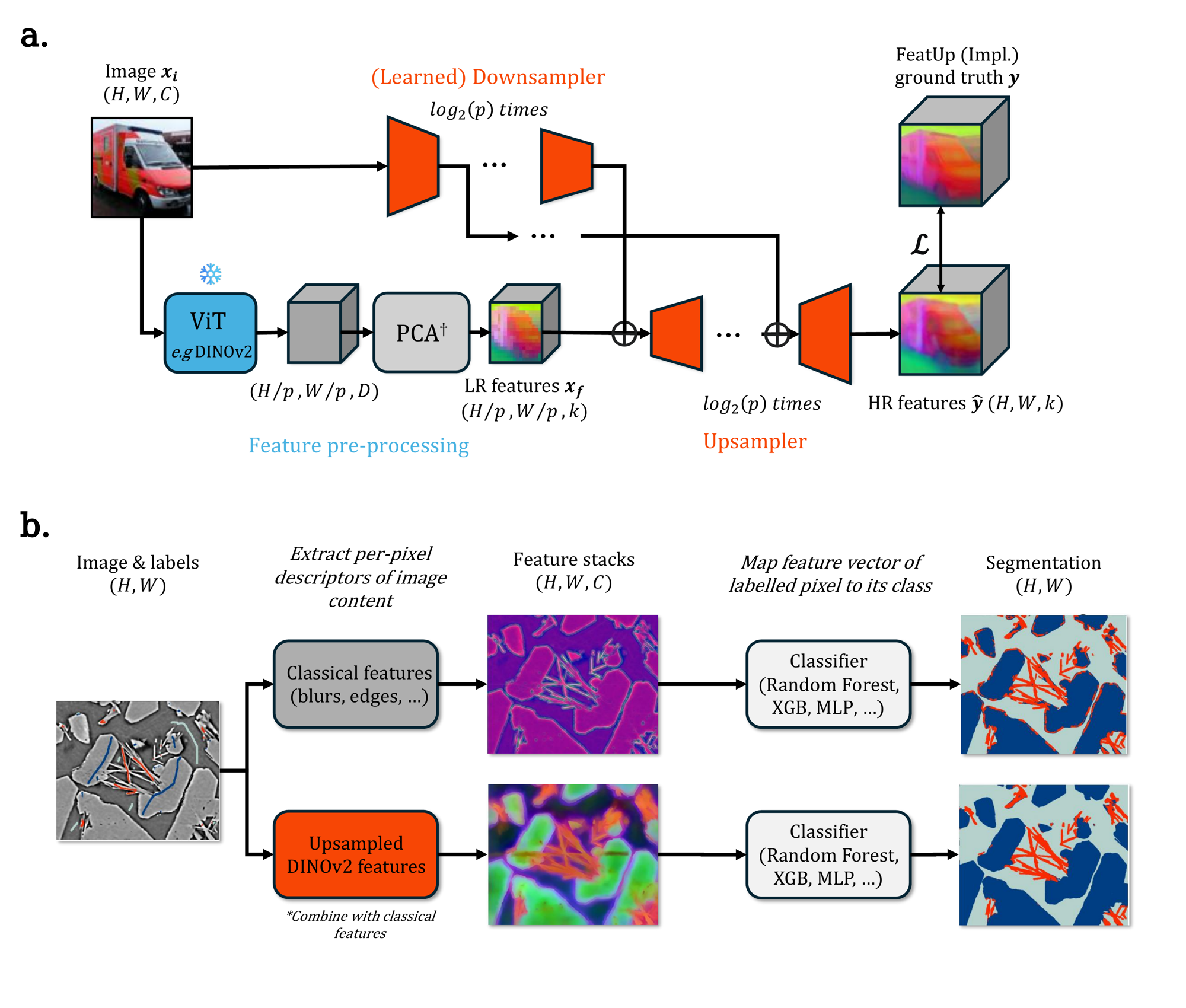}
    \caption{\textbf{a.} Model diagram for our upsampler. The input image $\bf{x_{i}}$ has its low-resolution (LR) features $\bf{x_{f}}$ computed from the ViT and pre-processed into a `FeatUp compatible' form. It is downsampled $log_{2}(p)$ times via a learned convolutional downsampler, whose activations are used as guidance for the convolutional feature upsampler. The result is compared to the FeatUp Implict high-resolution (HR) `ground truth' via a smooth MSE loss. The feature pre-processing and ground truth are computed offline and loaded during training. \textbf{b.} Once trained this upsampler can be used in interactive segmentation: low-resolution ViT features of micrographs can be computed, upsampled and combined with classical features before training a classifier to map from these features to the user-drawn labels (coloured lines on image). } 
    \label{fig:model_diagram}
\end{figure*}

Our approach is to learn a convolutional upsampler network to map from an image and its corresponding low-resolution features as extracted by a ViT to the high-resolution map produced by training an implict FeatUp network on that image. 
We use a convolutional network for its attractive time and memory cost with respect to input image size, and for its general suitability to image tasks. A diagram of the method can be seen in Figure \ref{fig:model_diagram}.

In this section we explain:
\begin{itemize}
    \item How we trained our upsampler to increase the resolution of the coarse ViT features with respect to the image, including the model architecture and the process of generating a set of high-resolution `ground-truth` features.
    \item The preprocessing steps (PCA) for the input low-resolutions features required by our choice of `ground-truth` during training, and the deep feature visualisations throughout the paper.
    \item How these upsampled ViT features are then used in interactive feature-based segmentation (\textit{i.e,} by concatenation with existing classical features). 
\end{itemize}

\subsection{Upsampler}
\label{sec:method:upsampler}

\subsubsection*{Choice of backbone}
Our approach works across a variety of ViT backbones, because it is trained on FeatUp implict ground-truths, which can be generated for various ViT and convolutional feature extractors \cite{FEATUP}.
DINO models with register tokens perform better \cite{REG_TOKENS}, as they have no anomalous background tokens with high attention, which can cause pathological behaviour in the FeatUp implict upsampler for some images.
We experiment using a fine-tuned DINOv2 checkpoint trained on 3D geometry, FiT3D\cite{FIT_3D}, which has sharper feature-maps and therefore may make it easier for our network to learn to upsample.

\subsubsection*{Dataset}
We train on a 2000 image subset of Imagenet Reduced, a set of 10,000 images taken from ImageNet \cite{IMAGENET}, a general-purpose classification dataset of natural images.
First, we generate low-resolution and high-resolution embedding pairs offline for each image by training a FeatUp implicit network offline. This is the longest step of the process, taking around three minutes per-image on a A100 GPU. Then, we train our upsampler to map from the image and its low-resolution embedding to the high-resolution `ground truth'.

Any generalist image dataset would work, and in future the upsampler could be trained on a larger dataset for better performance. A two-step training process could also improve performance (as in \cite{DINOv2}), where the network is initially trained on $256 \times 256$ images and later on larger, $512 \times 512$ images.

\subsubsection*{Architecture}
Our model has two components: first a learned convolutional downsampler extracts features from input image $\bf{x_{i}}$ (of size $H \times W$) iteratively, halving the resolution each time. These image features are used as guidance for a convolutional upsampler which takes as input the low-resolution ViT features $\bf{x_{f}}$, of dimension $(H/p, W/p, k)$. $p$ is the original patch size of the ViT and $k$ is the dimenion of the PCA from the FeatUp preprocessing, usually $k=128$ (see Section \ref{sec:method:feat_preprocess}).

$\bf{x_{f}}$ is upsampled $log_{2}(p)$ times until the features are at full resolution \textit{i.e,} $(H, W, k)$, and then compared via smooth MSE loss to the FeatUp implict `ground truth'. The layers of the down- and upsampler are taken from standard U-Net implementations; we note that the network acts similar to a U-Net, with the addition of the LR feature concatenation at the bottleneck \cite{U-NET}.
Further network architecture and hyperparameter details are available in Section \ref{sec:supp:hyperparams}.

\subsection{Feature preprocessing}
\label{sec:method:feat_preprocess}
When training their implicit upsampler, FeatUp prepares a dataset of features from a ViT of $N_t$ transforms of the original input image, then computes a shared PCA over these features (we call this `$\bf{x}_{\text{FU}, \text{lr}}$').
This shared PCA is then used to project the features of the original, untransformed image.
This is part of their NeRF analogy - "that multiview consistency of low-resolution signals can supervise the construction of high-resolution signals". 
Crucially, this shared PCA changes both the number of channels and the distribution of features when compared to the features produced by a ViT for the untransformed image in both low and high-resolution. This change is sometimes small when visualised (see Sections \ref{sec:method:feat_vis} and \ref{sec:supp:feat_prep_explained}), but is often significant.

The full dataset for the implicit network consists of $3000$ transformations of the image, but a PCA over $N_t=50$ is a good approximation (\textit{i.e,} the low- and high-res appear similar when visualised).
We found difficulty in training an upsampler end-to-end when taking the original ViT features and upsampling them with respect to the FeatUp implicit features due to this distribution shift - see Section \ref{sec:supp:feat_prep_explained}. We also attempted to train a `feature-transfer' network to map from the original ViT features to $\bf{x}_{\text{FU}, \text{lr}}$, but were unsuccessful.

To avoid this problem, we preprocess our low-resolution ViT features using this shared PCA before upsampling. This is done offline during training, alongside the high-resolution feature creation, and is computed and applied before upsampling at test-time. This means the output features from the model have implicitly had a PCA applied to them (the first feature channels capture the most variance), and that given a dataset of low- and high-resolution features of $k$ dimensions, compressed upsamplers of dimension $j < k$ can be trained by simply taking the first $j$ features, useful for GPU memory-constrained environments.

\subsubsection*{Feature Visualisation}
\label{sec:method:feat_vis}
As in previous works, we visualise our high dimensional features by taking a PCA over the features from $k=128$ or $d=384$ dimensions down to 3, and plotting the result as RGB channels, such that similar colours indicate similar features \cite{DINOv2, STRIDED}.
We do not need to train a PCA for FeatUp style features (see Section \ref{sec:method:feat_preprocess}). This visualisation, whilst useful, does not explain the full picture, especially for classical features where there are more edge-based filters. 
This means edge features explain more variance and so are highlighted, but it does not mean colour information (from Gaussian blurs \textit{etc.}) is absent, merely unemphasised. 

\subsection{Interactive segmentation workflow}
Following a similar setup to previous work \cite{WEKA_PAPER, ILASTIK, HR-DV2}, we extract and concatenate our deep features to the classical ones before training a classifier to map to the user labels.
We focus on classifiers which train quickly, like random-forests\cite{RANDOM_FORESTS}, XGBoost\cite{XGBOOST} and logistic regression models.
This, combined with the efficient nature of our feature upsampler, allows quick, real-time interactive segmentation (see Section \ref{sec:results:perf_landscape}).

Explicitly, the classical process trains a pixelwise classifier $f_{\theta}: \mathbb{R}^{N} \rightarrow \mathbb{Z}, \mathbb{Z} \in [1, ..., C]$ over the set of features of all labelled pixels where $N$ is the dimension of the vector produced using the `classical' image processing operations and $C$ is the number of classes. We concatenate the $k$-dimensional (upsampled) deep features for each pixel onto the $N$-dimensional classical features and train a classifier $f'_\theta$, $f'_{\theta}: \mathbb{R}^{N+k} \rightarrow \mathbb{Z}, \mathbb{Z} \in [1, ..., C]$ - we denote this as `+ HR ViT features' or `+HR ViT'. 

`Classical' refers to a Python reimplementation of the Weka featurisation process described in Section \ref{sec:theory:trainable_seg}, which we use as a baseline comparison.
The same labels and classifier type were used for all `Classical' and `+ HR ViT features' comparisons.
Unless otherwise stated, we use our convolutional upsampler to upsample DINOv2-S features following Sections \ref{sec:method:upsampler} \& \ref{sec:method:feat_preprocess}, and an XGBoost classifier (with default hyperparameters) to map from features to labels.

We note no pre-processing (denoising, histogram equalization, \textit{etc}) or post-processing (\textit{e.g,} modal filters, small hole removal, \textit{etc}) were used, though the approach is of course compatible with them. 

\subsection{Evaluation metric}
For evaluating segmentation performance in Section \ref{sec:results} we use mean Intersection over Union (mIoU).  The mIoU (for a binary segmentation) measures the ratio of true positives to true positives plus false positives (\textit{i.e,} the positive predictive value) across all the pixels in a prediction when compared to a ground truth.
This can then be applied to multi-class segmentation by treating the predictions for each class as a binary segmentation and averaging over them. 

\section{Results \& discussion}
\label{sec:results}

\subsubsection*{Performance landscape}
\label{sec:results:perf_landscape}
\begin{figure*}
\centering
    \includegraphics[width=\linewidth]{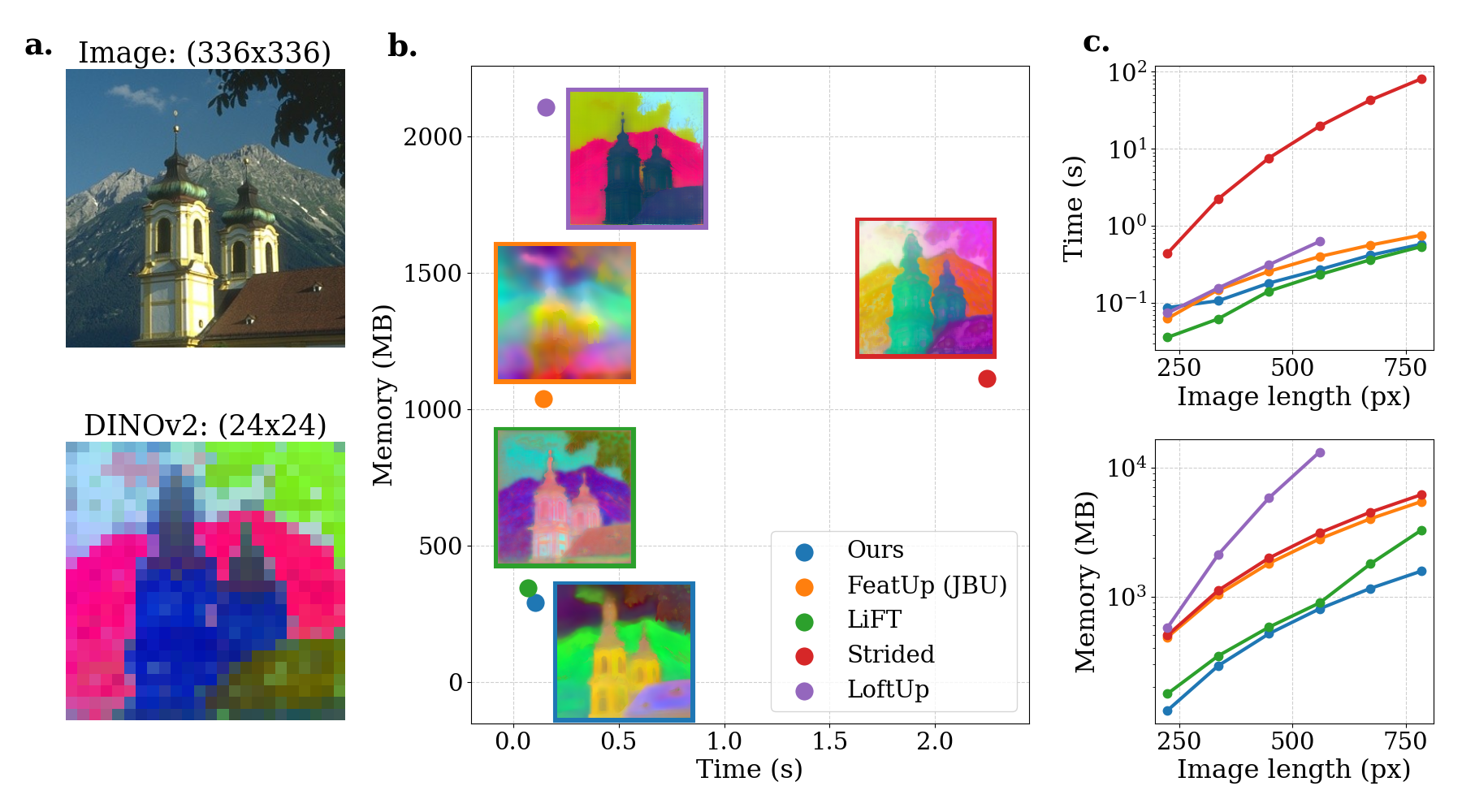}
    \caption{Performance landscape of various ViT feature upsamplers. \textbf{a.} the input image and corresponding patch-based feature vectors extracted by DINOv2. All feature vectors have been reduced to 3 dimensions via PCA and plotted as RGB channels. \textbf{b.} memory, time and feature-quality comparison for input image at size (336, 336); bottom-left is better. \textbf{c.} time and memory scaling as image side length is increased. `LiFT' uses DINO (not DINOv2) features, `Ours' and `FeatUp JBU' use a modified feature space (seee Section \ref{sec:method:feat_preprocess}). Metrics are measured end-to-end \textit{i.e,} extraction plus upsampling.}
    \label{fig:perf_landscape}
\end{figure*}

To be used successfully in interactive segmentation, it must be quick and cheap to compute the high-resolution deep features, and these features should be well-localised/sharp. Therefore, we compare the time and memory costs of the various generalised upsampling approaches (strided, JBU, LiFT, LoftUp, ours) in Figure \ref{fig:perf_landscape}, plotting them alongside an example feature-map for a reference image. More qualitative examples of feature map comparisons for micrographs can be found in Section \ref{sec:supp:choice_of_upsampler_on_is}.

We find that our approach and LiFT have the most favourable time \& memory scaling (strided has high time cost; JBU and LoftUp a high memory cost), and similar feature sharpness. Our approach has more focus on semantics rather than colour information, but this is because it is trained on DINOv2 whereas LiFT is trained on DINO.

The ability to train a `compressed' upsampler for $k=16$ features (see Section \ref{sec:method:feat_preprocess}) allows us to process relatively large ($\sim 2000 \times 1000$ pixels) images on a laptop GPU (4GB NVIDIA RTX 3050M), as the cost of storing a $(H,W, 64/128/384)$ in memory or on disk scales rapidly. This improved scaling, and the efficiency of our approach in general, is important for microscopists, whose images can often be hundreds of megapixels.

\subsection{Interactive segmentation}
Following our previous work\cite{HR-DV2}, we concatenate these upsampled deep features with the classical features to perform interactive segmentation for a variety of material micrographs.
We do this for large micrographs displaying complex phases, as well as testing their generalisation ability on predefined benchmark datasets.

In both the examples and the benchmarks we find that adding the HR ViT features offers the classifier additional relevant information on which it can condition its predictions - we can see evidence of this in the feature map visualisations in Figure \ref{fig:intro_column} and in Section \ref{sec:supp:feature_vis_micrographs}. Despite not being trained on micrographs, the ViT models have learned useful primitives that generalise, like foreground vs background, interiority, (perceived) depth, shape \textit{etc.}

\subsubsection*{Single-image examples}
\label{sec:results:is_examples}

\begin{figure*}
\centering
    \includegraphics[width=\linewidth]{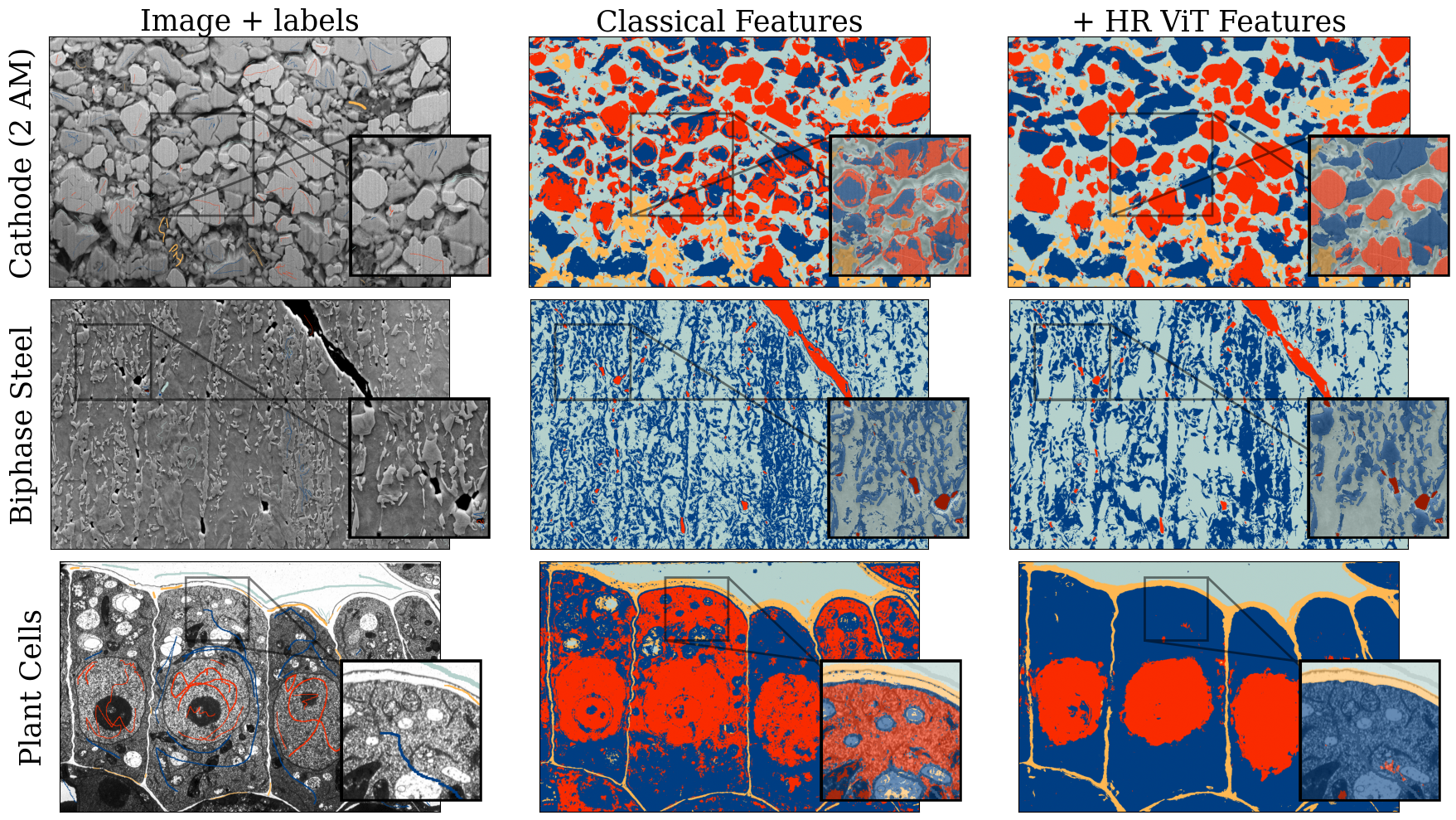}
    \caption{Interactive segmentation results on SEM micrographs using `Classical' features and `Classical + HR ViT features', which are features from DINOv2-S which have been upsampled using our model. The labels and classifier type were the same across each example. The addition of the HR ViT features admits a much improved segmentation for each material: the in/out- of plane material in the cathode\cite{BIPHASE_CATHODE}, the ferrite/martensite distinction in dual-phase steel\cite{BIPHASE_STEEL}, and the nucleus vs. cytoplasm and organelles of the plant cells\cite{PLANT_SEM}. The featurisation, upsampling and classifier training can be completed in under 10s on a laptop GPU. Labels may be single-pixel thin and therefore hard to see. }
    \label{fig:is_examples}
\end{figure*}

Figure \ref{fig:is_examples} shows a comparison between the segmentations produced using the `Classical' features and the `+ HR ViT features' for a series of scanning electron microcopy (SEM) micrographs. 
The first is an SEM of a battery cathode with four phases\cite{BIPHASE_CATHODE}: pore (and out-of-plane material), carbon binder, and two different active materials (light and dark grey particles).
The ViT features allow a much better distinction between material in (\textit{i.e,} flat, usually curtained) and behind the imaging plane. 
This distinction is important for accurately determining a battery's porosity for use in later physics-based simulations\cite{KINTSUGI}, and for understanding the impact of processing conditions like calendaring.

\begin{figure*}
\centering
    \includegraphics[width=\linewidth]{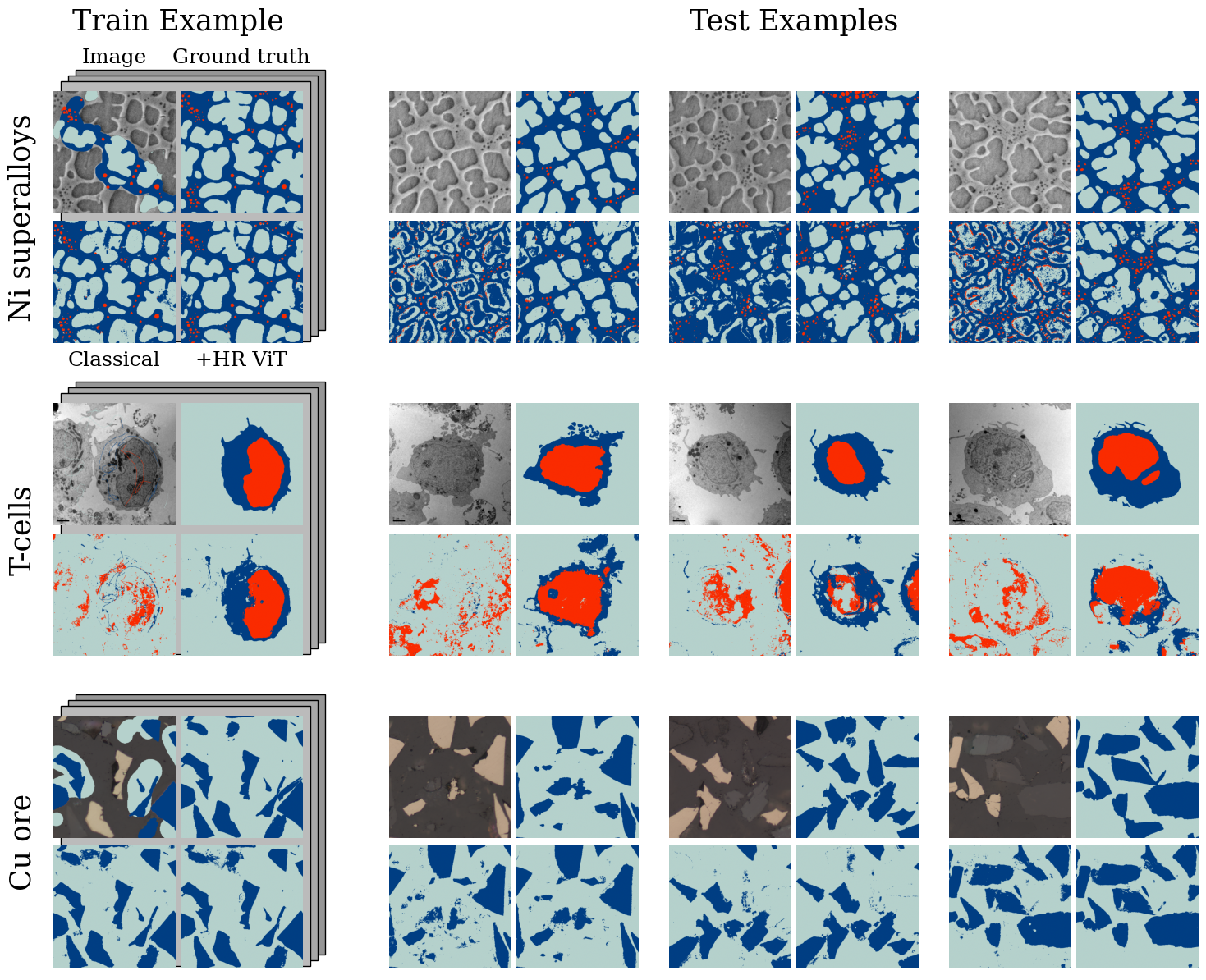}
    \caption{Generalisation results of classifiers trained on classical and +HR ViT features over three datasets: SEM images of Nickel superalloys, TEM images of human T-cells and reflected light microscopy of copper ore in resin. The classifiers are trained on labels and features of four images and applied unseen to the rest. +HR ViT features (bottom left of each panel) allow more accurate segmentations relative to the ground truth (top right). Full results are available in Table \ref{tab:is_benchmark}.  }
    \label{fig:is_benchamrk}
\end{figure*}

The second micrograph is dual-phase steel, with ferrite, martensite (raised) and some damage (black voids)\cite{BIPHASE_CATHODE}. 
We see that the classical features struggle to consistently distinguish the centres of large martensite regions and the flat ferrite regions. 
The third is an SEM of plant cells\cite{PLANT_SEM}, where the four phases are the background, cell wall, cytoplasm and nucleus; the classical features struggle to differentiate between the nucleus and the rest of the cells due to similar intensities.

Each of these images was roughly $2000 \times 1000$ pixels large, and was segmented on a laptop GPU on a custom interactive segmentation app. Although our upsampler is most suited to this specific environment, we note that the other upsamplers could be used, were more memory available (FeatUp JBU, LiFT). 
We study the impact of the ViT feature upsampler on interactive segmentation in Section \ref{sec:supp:choice_of_upsampler_on_is}, finding that the simple approaches (bilinear upsampling, strided) cause artefacts.

\subsection*{On benchmark datasets}
\label{sec:results:is_benchmarks}
Few labelled interactive segmentation datasets exist for materials science, and there are fewer `benchmark datasets' which include examples across different material systems and instruments.
We collate a series of three datasets of micrographs and their associated ground truth segmentations into a modest benchmark which aims to test the accuracy and generalizability of the feature-sets. 

The first dataset, `Ni', is a set of SEM micrographs of nickel superalloys with tertiary precipitates, taken from ref \cite{MICRONET_CNN}. The second, `T-cell', is series of transmission electron microscopy (TEM) images of Jurkat T-cells \cite{T-CELLS} with three classes: background, nucleus, cytoplasm. The third dataset `Cu ore' is reflected light microscopy (RLM) images of copper ore in epoxy resin \cite{CU_ORE}, with two classes: ore and resin (background). 

For each of these datasets we train using the features and labels of four exemplar images and apply the classifiers unseen to the other images in the dataset (18, 26 and 26 respectively). We measure the mIoU of the prediction relative to the ground truth, and report the results in Table \ref{tab:is_benchmark}. 
Example train and test data, as well as classifier predictions, are available in Figure \ref{fig:is_benchamrk}. The mIoU is measured across all images (train included) - this is justified as the labels are themselves sparse (\textit{i.e,} do not cover the whole image) and are consistent across feature-sets. 
Labels were produced using the SAMBA web-app's smart labelling tool\cite{SAMBA}.

\begin{table}[H]
    \centering
    \begin{tabular}{|c| c |c c|}
    \hline
    \textbf{Dataset} & \textbf{${N_\text{labels}}$}& \multicolumn{2}{c|}{\textbf{Class-avg mIoU}}  \\
    & & Classical & +HR ViT \\
    \hline
    \multirow{2}{*}{Ni} & 4 & 0.56 $\pm$ 0.16 & \textbf{0.75 $\pm$ 0.11 }\\
     & 22 & 0.73 $\pm$ 0.16 & \textbf{0.82 $\pm$ 0.10 }\\
    \multirow{2}{*}{T-cell}  & 4 & 0.34 $\pm$ 0.09 & \textbf{0.64 $\pm$ 0.16 }\\
     & 22 & 0.41 $\pm$ 0.08 & \textbf{0.80 $\pm$ 0.13  }\\
    \multirow{2}{*}{Cu ore}  & 4 & 0.85 $\pm$ 0.06 & \textbf{0.87 $\pm$ 0.03 }\\
     & 22 & 0.88 $\pm$ 0.05 & \textbf{0.89 $\pm$ 0.03 } \\
    \hline
    \end{tabular}
    \caption{Average mIoU of the three classes relative to the ground truth for three example benchmark datasets: Ni superalloys\cite{MICRONET_CNN}, human T-cells\cite{T-CELLS} and Cu ore\cite{CU_ORE} - in both 4- and fully-labelled images. Standard deviation is measured across images in the dataset. }
    \label{tab:is_benchmark}
\end{table}

In agreement with previous observations, we see that the ViT features allow more accurate segmentations, increasing the mIoU (sometimes greatly) across all datasets. For the Ni superalloys, this means more consistency across large exposure variations, especially of the tertiary precipitates. On the T-cells this manifests as better distinction between foreground/background cells, and of the interior nucleus.

\begin{figure}[H]
\centering
    \includegraphics[width=0.9\linewidth]{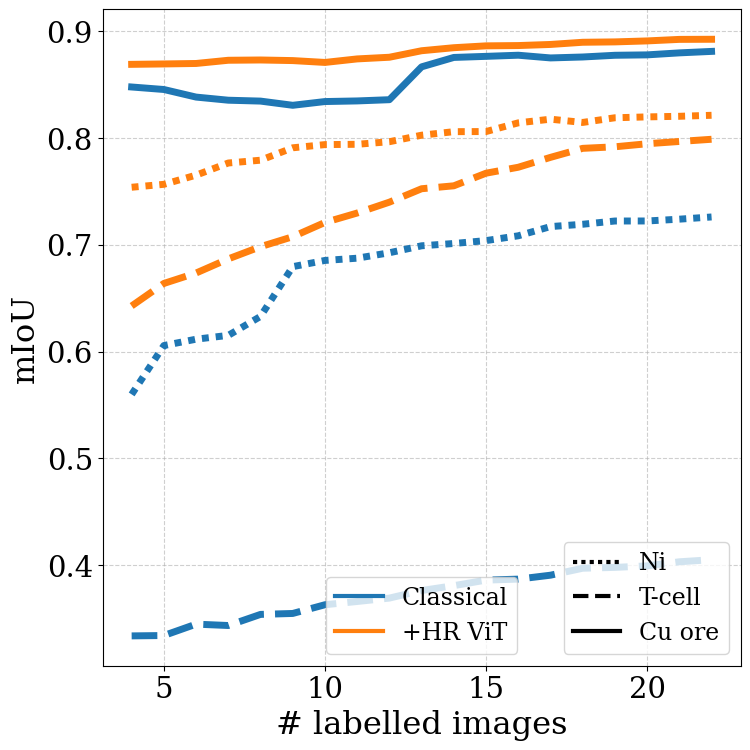}
    \caption{Class-averaged mIoU as function of number of labelled images for the various benchmark datasets for each feature-set. We see the +HR ViT features outperform the classical feature-set for each dataset.}
    \label{fig:is_longlimit}
\end{figure}

Finally, we study the long limit on these benchmark datasets, increasing the number of sparsely-labelled images we train the classifiers over, and measuring the mIoU over all images each time. We present the results in Figure \ref{fig:is_longlimit} (and the highest achieved value in Table \ref{tab:is_benchmark}), and see that the deep features scale well with added training data, achieving high mIoUs. 

Some sharp increases in performance for the classical feature-set can be seen, this is interpretable as a data-ordering problem, where some useful feature or ambiguity becomes labelled and performance across the rest of the images increases. This ordering however does not affect the result in the long limit.

Example predictions for these `fully-trained' classifiers on the Ni superalloys is available in Figure \ref{fig:supp:max_label_preds}, we note that the predictions with +HR ViT features align much more closely with the ground truth. 
For some datasets (Ni, T-Cell) it took five to six times longer to train classifiers on the classical features than the +HR ViT features - the trees had to be grown to a larger depth to make adequate splits. This time increase was not seen for the Cu ore dataset, where the classical and +HR ViT feature-sets had more similar performance.

It is unsurprising that the addition of the HR ViT features improves segmentation performance - all the classical feature-set information is still present - but the increase in perceived quality and mIoU is marked, as is the speed increase during classifier training. We demonstrate this improvement is not attributable to regularization (\textit{i.e,} by comparing to the addition of random or uninformative channels) in Section \ref{sec:supp:perf_justification} - and find the +HR ViT features must be supplying new, useful information to the classifier.

\subsection*{Explicit comparison to CNNs}
\label{sec:results:cnn_comparison}

\begin{figure*}
\centering
    \includegraphics[width=\linewidth]{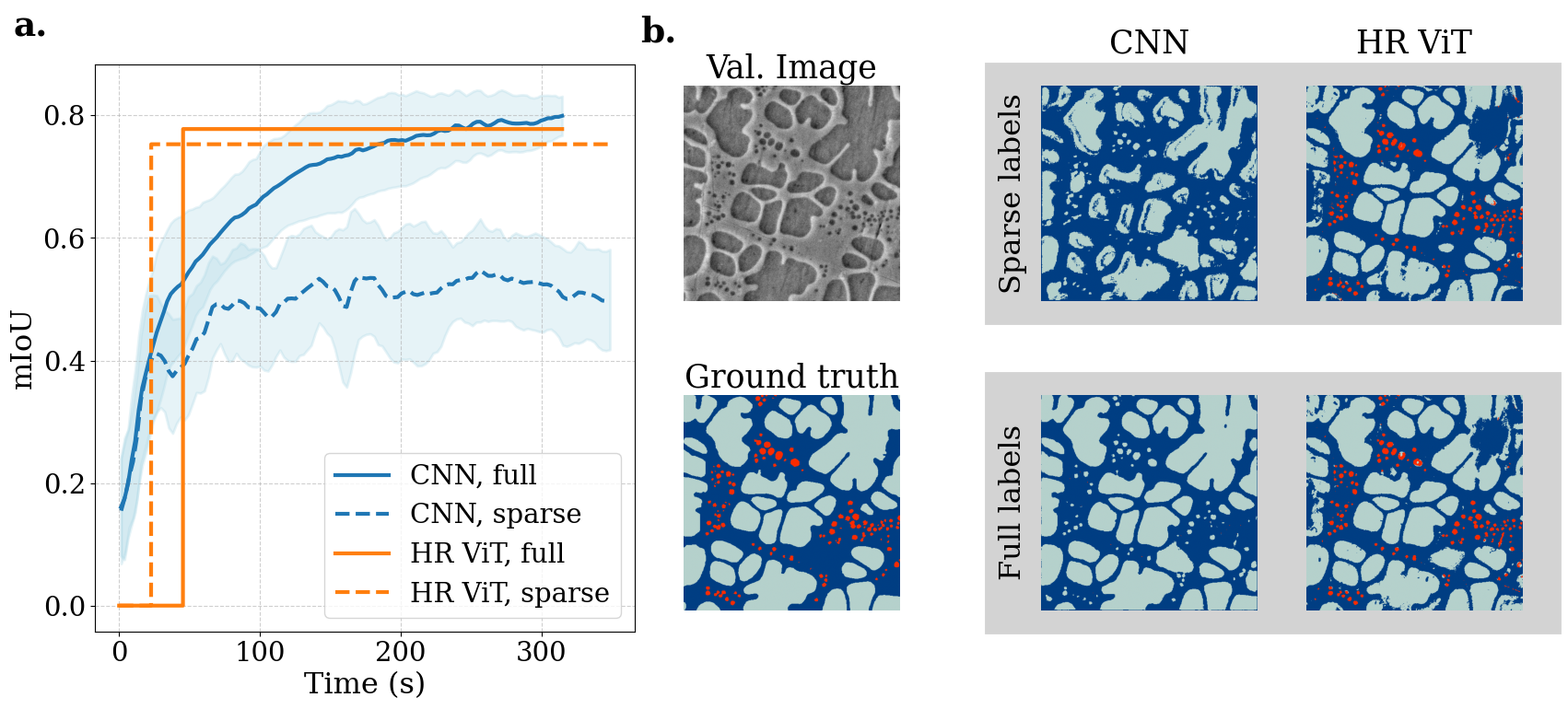}
    \caption{Comparison of segmentation performance for a CNN vs. interactive segmentation supplemented with upsampled deep features (HR ViT) on the 3-phase Ni superalloy dataset. Four micrographs were used for training, either with `sparse' labels from Section \ref{sec:results:is_benchmarks} or `full' labels using the whole ground truth. For the CNN values have been averaged over 10 training runs.  \textbf{a.} mIoU as a function of time for the approaches, we see that interactive segmentation can rapidly achieve high mIoUs, even with sparse labels. \textbf{b.} performance on a validation image once our interactive segmentation approach has finished segmenting (featurising + training + applying).}
    \label{fig:cnn_comparison}
\end{figure*}

In order to justify our claim that `you may not need a U-net', we compare our approach to a recent micrograph-pretrained CNN model, `MicroNet' \cite{MICRONET_CNN}.
MicroNet is a U-Net architecture with a series of backbones that have been trained on ImageNet \cite{IMAGENET} and then on a proprietary large micrograph dataset.
It is then fine-tuned using task-specific image and segmentation mask pairs - we view this large backbone + task-specific training as similar to the (deep) feature extraction + classifier training of our approach.

We compare the segmentation quality (mIoU) over all 22 images as a function of time for MicroNet and our approach in Figure \ref{fig:cnn_comparison}. Both approaches were trained using four micrograph - segmentation pairs from the Ni-superalloy dataset introduced by MicroNet. We either train on the sparse labels from Section \ref{sec:results:is_benchmarks} ($\sim 20\%$ pixels labelled), or using the full ground truth segmentations. For MicroNet time is measured over each epoch of training, and for HR ViT time is measured for featurisation + classifier training + classifier application for all 22 images (\textit{i.e,} a conservative estimate for our approach). 

To train MicroNet on sparse labels we mask its standard Binary Cross Entropy loss to only the labelled pixels. Other parameters are kept the same (learning rate, number of epochs \textit{etc.}), consistent across the different labelling methods (full vs. sparse). The U-Net++ configuration with a ResNet50 encoder \cite{RESNET50} was used for MicroNet.

Our approach reaches a high mIoU quickly, even with sparse labels (75\% in 20.4s for sparse labels, 79\% in 38s for full labels). We note this was achieved using CPU XGBoost and a 128-channel upsampler - there is room for further speedup. MicroNet takes more time, taking around 4 minutes to reach 80\% mIoU with full labels, and struggles to train well with sparse labels. 
This indicates our approach is markedly more time and data efficient than using a CNN, useful if fully-labelled ground truths are unavailable or costly to obtain (fully-labelling large micrographs can take days\cite{ACTIVE_LEARNING_BATTERY_SEG}). 

However, given longer time we expect MicroNet (with fully-labelled data) would outperform HR ViT, which we observe in Figure \ref{fig:cnn_comparison}. This is not unexpected: fine-tuning feature extraction on task-specific data should lead to the best performance.

We stress that the featurisation step for +HR ViT need only be computed once, and further labels can be added to quickly retrain and improve the classifier, for example on other images in the validation set. This compounds the advantage over the CNN: mistakes can be corrected interactively using sparse labels well before the CNN has finished training.

\section{Conclusions \& future work}
\label{sec:conclusion} 

To conclude, we have introduced a fast convolutional feature upsampler for ViT features and demonstrated the capacity of these features to greatly improve interactive micrograph segmentation. We do this for a range of materials, instruments and image resolutions, in resource constrained environments all in matter of seconds. This approach is flexible, outperforms CNNs, and requires far less labelled training data. 

There is scope for future work to improve various aspects of the approach: reducing or removing the positional bias implicit in the ViT features (see Section \ref{sec:supp:limit_pos_bias}), finding ways of improving the feature-preprocessing step and ensuring consistency of ViT features (\textit{i.e,} in a volumetric stack). It is also important to account for the fact that the majority of micrographs are greyscale, and that feature foundation models (and our upsampler) are trained on natural, RGB images. Training a feature foundation model on micrographs could ameliorate this, as well as allowing better expressivity of microstructural features likes grains, cracks and pores.

We hope this work will encourage further integration (and clever adaptation) of foundation model features into materials imaging workflows, like segmentation, denoising or property prediction.

\section*{Code Availability}
\label{sec:code_availability}
The Python reimplementation of the classical featurisation process is available \href{https://github.com/tldr-group/interactive-seg-backend}{here} and the GUI for the user app is available \href{https://github.com/tldr-group/interactive-seg-gui}{here}. The code for the ViT feature upsampler and to reproduce the paper figures, alongside instructions on how to install these three packages together, is available \href{https://github.com/tldr-group/vulture}{here}.

\section*{Data Availability}
\label{sec:data_availability}
 The data used in this study (micrographs, labels) are available on \href{https://zenodo.org/records/16993498}{zenodo}.

\section*{Acknowledgements}
This work was supported by funding from the the EPRSC and SFI Centre for Doctoral Training in Advanced Characterisation of Materials (EP/S023259/1 received by RD) and the Royal Society (IF\textbackslash R2\textbackslash 222059 received by AV as a Royal Society Industry Fellow).

The authors would like to thank other members of the TLDR group for their testing and feedback.

\section*{Authorship}
RD conceived the concept, wrote the code and performed analysis for the manuscript. AV and SJC contributed to the development of the concepts presented in all sections of this work and made substantial revisions and edits to all sections of the draft manuscript.

\section*{Competing interests}
The authors declare no competing interests.

\section*{References}
\addcontentsline{toc}{section}{References}

\def\addvspace#1{}

	\renewcommand{\refname}{ \vspace{-\baselineskip}\vspace{-1.1mm} }
	\bibliographystyle{ieeetr}
    \bibliography{main}

\end{multicols}

\newpage
\section*{Supplementary}
\input{supp}

\end{document}

%% file: styles_and_packages/preamble.tex
\usepackage{geometry}
\usepackage{titling}
\usepackage[utf8]{inputenc}
\DeclareUnicodeCharacter{2212}{-}
\usepackage{amsmath}
\usepackage{siunitx}
\usepackage{lineno}
\usepackage{authblk}
\usepackage{lscape}
\usepackage{graphicx}
\usepackage{subcaption}
\usepackage[font={footnotesize},labelfont=bf]{caption}
\usepackage[super,numbers,sort&compress]{natbib}

\usepackage{gensymb}
\usepackage{setspace}
\usepackage{multicol}

\usepackage{afterpage}
\usepackage{xtab}
\usepackage{amssymb}
\usepackage{rotating}
\usepackage{xr}
\usepackage[hidelinks]{hyperref}
\hypersetup{
    colorlinks=true,
    linkcolor=black,
    filecolor=magenta,      
    urlcolor=cyan,
    }

\usepackage{cleveref}
\usepackage{algorithm}
\usepackage[noend]{algpseudocode}
\usepackage{enumitem}
\setlist[itemize]{leftmargin=*}
\usepackage{float}
\usepackage{multirow}

\usepackage{styles_and_packages/arxiv}

\usepackage{dirtytalk}
\usepackage{txfonts}
\usepackage{soul}

%% file: supp.tex
\setcounter{section}{0}
\setcounter{figure}{0}
\setcounter{table}{0}
\renewcommand*{\theHsection}{S.\the\value{section}}

\makeatletter
\renewcommand \thesection{S\@arabic\c@section}
\renewcommand\thetable{S\@arabic\c@table}
\renewcommand \thefigure{S\@arabic\c@figure}
\makeatother

\section{Classical feature-set}
\label{sec:supp:classical_feats}

We used a Python reimplementation (using numpy\cite{NUMPY} and scikit-image\cite{SCIKIT_IMAGE}) of the default Weka\cite{WEKA_PAPER} feature-set for the classical features.
\begin{enumerate}
  \item \textbf{Gaussian blurs} at each pixel for a set of strengths $\sigma \in \{0, 1, 2, 4, 8, 16 \}$. \label{supp:li:feats:gauss}
  \item \textbf{Sobel edge detection} at each pixel for each of the Gaussian filtered arrays in \ref{supp:li:feats:gauss}.
  \item \textbf{Hessian texture filter} at each pixel for each of the Gaussian filtered arrays in \ref{supp:li:feats:gauss}, extracting the first two eigenvalues, as well as the mod, trace and determinant of the Hessian matrix.
  \item \textbf{Difference of Gaussians} of each of the Gaussian arrays in \ref{supp:li:feats:gauss}.
  \item  \textbf{Membrane projections:} convolution of the image array with a stack of line kernels oriented at $30\degree$ angle increments in $[0\degree, 180\degree).$
\end{enumerate}

\section{Hyperparameters}
\label{sec:supp:hyperparams}
\begin{figure*}
\centering
    \includegraphics[width=0.9\linewidth]{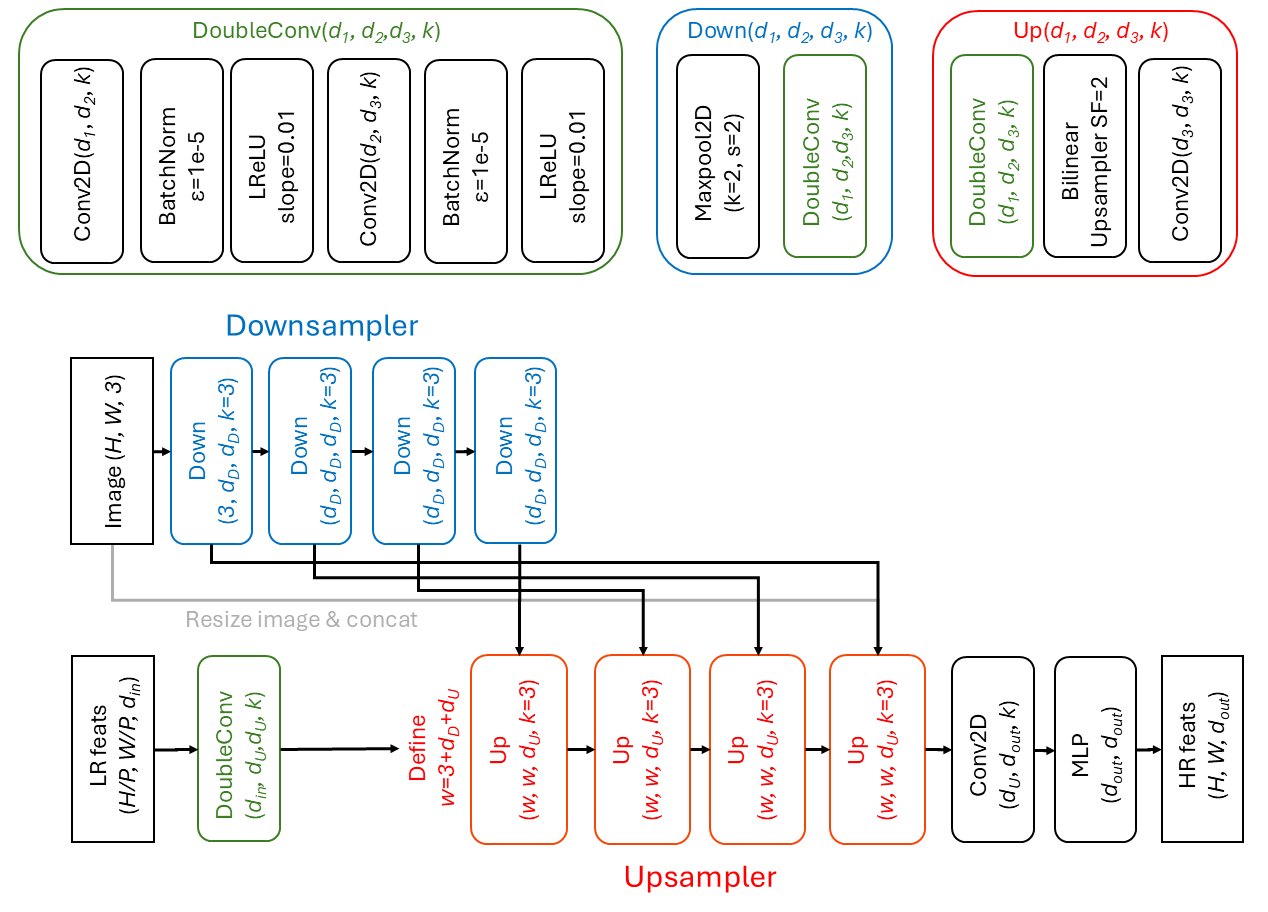}
    \caption{Model architecture. `DoubleConv', `Down' and `Up' are standard U-Net layers, with standard parameters. The image (and corresponding features computed from the downsampler) are concatenated with the DINOv2 features at each layer of the upsampler as guidance. } 
    \label{fig:supp:model_arch}
\end{figure*}

The training hyperparameters for our upsampler is detailed in Table \ref{tab:hyperparms}. The network architecture is shown in Figure \ref{fig:supp:model_arch}.

\begin{table}
    \centering
    \begin{tabular}{|c | c | c | c|}
    \hline
    \textbf{Input channels, $d_{in}$} & $32 / 64 / 128 / 384$ & \textbf{Output channels, $d_{out}$} & $32 / 64 / 128 / 384$ \\
    \textbf{Hidden channels, $d_{U}$} & $32 / 64 / 128 / 96$ & \textbf{Downsampler channels, $d_{D}$} & $32 / 32 / 64 / 64$ \\
    \textbf{Kernel size, $k$} & 3 & \textbf{Augmentations} & flip\_h, flip\_v, rotate $(0, 90, 180, 270)$ \\
    \textbf{Normalization} & L1 & \textbf{Loss} & Smooth L1 \\
    \textbf{Optimizer} &  AdamW & \textbf{Learning rate} & 1e-4 \\
    \textbf{Batch size} & 32 & \textbf{Epochs} & 5000 \\

    \hline
    \end{tabular}
    \caption{Training hyperparameters for the feature upsampler. }
    \label{tab:hyperparms}
\end{table}

\section{Feature preprocessing explained}
\label{sec:supp:feat_prep_explained}
FeatUp computes and applies a $k$-component PCA during training of their per-image implicit upsampler which we use as ground-truth high-resolution training targets. This PCA is computed over the features of 50 transformations of an input image. These transformations are randomly sampled and consist of flips, pads and zooms. This PCA not only changes the number of channels available in the feature-stack, but also shifts the feature distribution. 

\begin{figure*}
\centering
    \includegraphics[width=\linewidth]{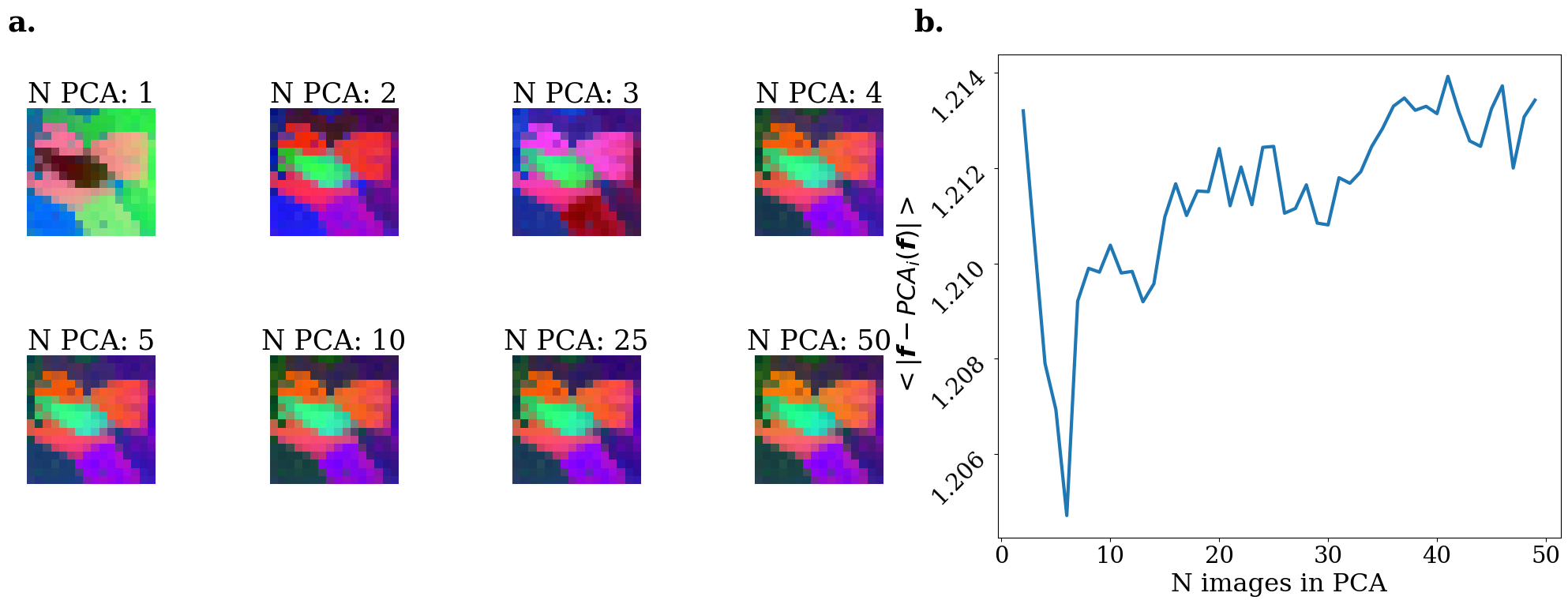}
    \caption{\textbf{a.} visualisation of the 128-component PCA across the DINOv2 features of N random transformations of the same image. Note N=1 is the PCA of the untransformed image. \textbf{b.} patchwise MAE of N random image PCA of features vs 1 image (untransformed) PCA of features, averaged over 100 images. Small numbers of transformations (N=2, 3) have a large impact on the subspace and therefore on the  difference from the reference image - this difference then increases as more images added, \textit{i.e,} the feature distribution shifts further.} 
    \label{fig:supp:dist_shift}
\end{figure*}

We present an example of this distribution shift in Figure \ref{fig:supp:dist_shift} \textbf{a.}, comparing the 128-component PCA of the untransformed image to subsequent PCAs with features of random transformations of the image included (\textit{i.e,} the FeatUp PCA process). The distribution shift can be seen in the colour changes of the patch features.
In  Figure \ref{fig:supp:dist_shift} \textbf{b.} we show the average patchwise MAE from the N-image PCA to the 1-image PCA, averaged over 100 images.

This data distribution shift made it more difficult to learn a lightweight upsampler from DINOv2 features to high-resolution FeatUp ground truths (which were implicitly PCA'd), as it had to learn both a complex distribution shift and the upsampling task. To fix this, we applied this feature preprocessing (\textit{i.e,} projection with a PCA fit over the features of 50 transformations of the input image) to the low-resolution DINOv2 features before upsampling. We must therefore apply this preprocessing during inference on new images.


\section{Impact of upsampler on interactive segmentation}
\label{sec:supp:choice_of_upsampler_on_is}

\begin{figure*}
\centering
    \includegraphics[width=0.9\linewidth]{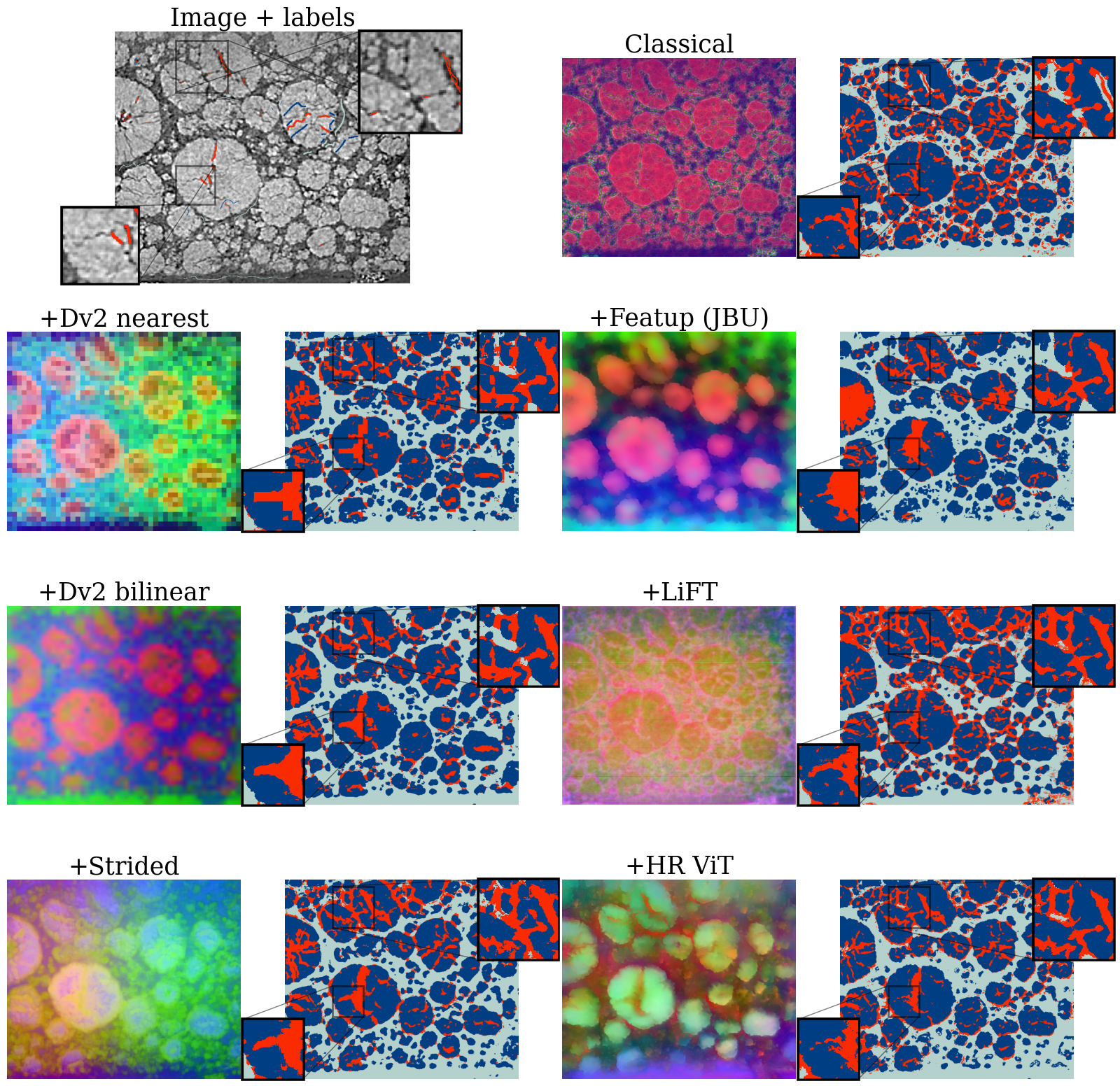}
    \caption{Impact of DINOv2 feature upsampler on interactive segmentation for a slice of an XCT of an NMC cathode. Note simple upsampling schemes (nearest-neighbour, bilinear) cause artefacts in the segmentations, as do other ViT feature upsamplers (FeatUp, LiFT).} 
    \label{fig:supp:upsampler_choice}
\end{figure*}

We compare different DINOv2 feature upsamplers when used in interactive segmentation in Figure \ref{fig:supp:upsampler_choice}. For the same image (XCT slice of a cracked NMC cathode\cite{NOISY_CATHODE_XCT}) and labels, we compute the classical features and concatenate them with the deep features from a given upsampler, train an XGB classifier to map features to labels and then apply to generate a segmentation.

We see the simple upsampling choices like nearest-neighbour upsampling and bilinear suffer from artefacting. Other approaches like FeatUp and Lift display either blurring or positional effects. 

Similar to \cite{HR-DV2, BENCHMARK_FU}, we find that bilinear feature upsampling is cheap and effective, especially when we combine it with classical features in interactive segmentation. The benefit of more complex upsamplers (ours, FeatUp, LoftUp, \textit{etc.}) comes when the deep features are needed to distinguish fine features (less than the patch size) like cracks inside or the edges of particles in Figure \ref{fig:supp:upsampler_choice}. 

This becomes more important as the patch size increases from 14px \cite{DINO} to 16px in DINOv3 \cite{DINOv3}. A side benefit of our approach is that compressed feature stacks ($k=16, 32, 64, 128$) are easier to store (in memory or on disk) than the full $k=384$ from bilinear upsampling.

\section{How many deep features do we need?}
\label{sec:supp:n_deep_feats}

\begin{table}[H]
    \centering
    \begin{tabular}{|l|c|}
    \hline
    \textbf{Number of added HR ViT channels} &  \textbf{Class-avg mIoU}  \\
    \hline
    0 & 0.53 $\pm$ 0.14 \\
    1 & 0.59 $\pm$ 0.15  \\
    16   & 0.72 $\pm$ 0.16   \\
    32  & 0.73 $\pm$ 0.16   \\
    64  & 0.74$ \pm$ 0.16 \\
    128  & \textbf{0.74 $\pm$ 0.16 }\\
    \hline
    \end{tabular}
    \caption{mIoU on the Ni superalloy dataset with 4 labelled examples as a function of number of added HR ViT features.}
    \label{tab:supp:how_many_channels}
\end{table}

\section{More feature visualisation examples}
\label{sec:supp:feature_vis_micrographs}
\begin{figure*}
\centering
    \includegraphics[width=\linewidth]{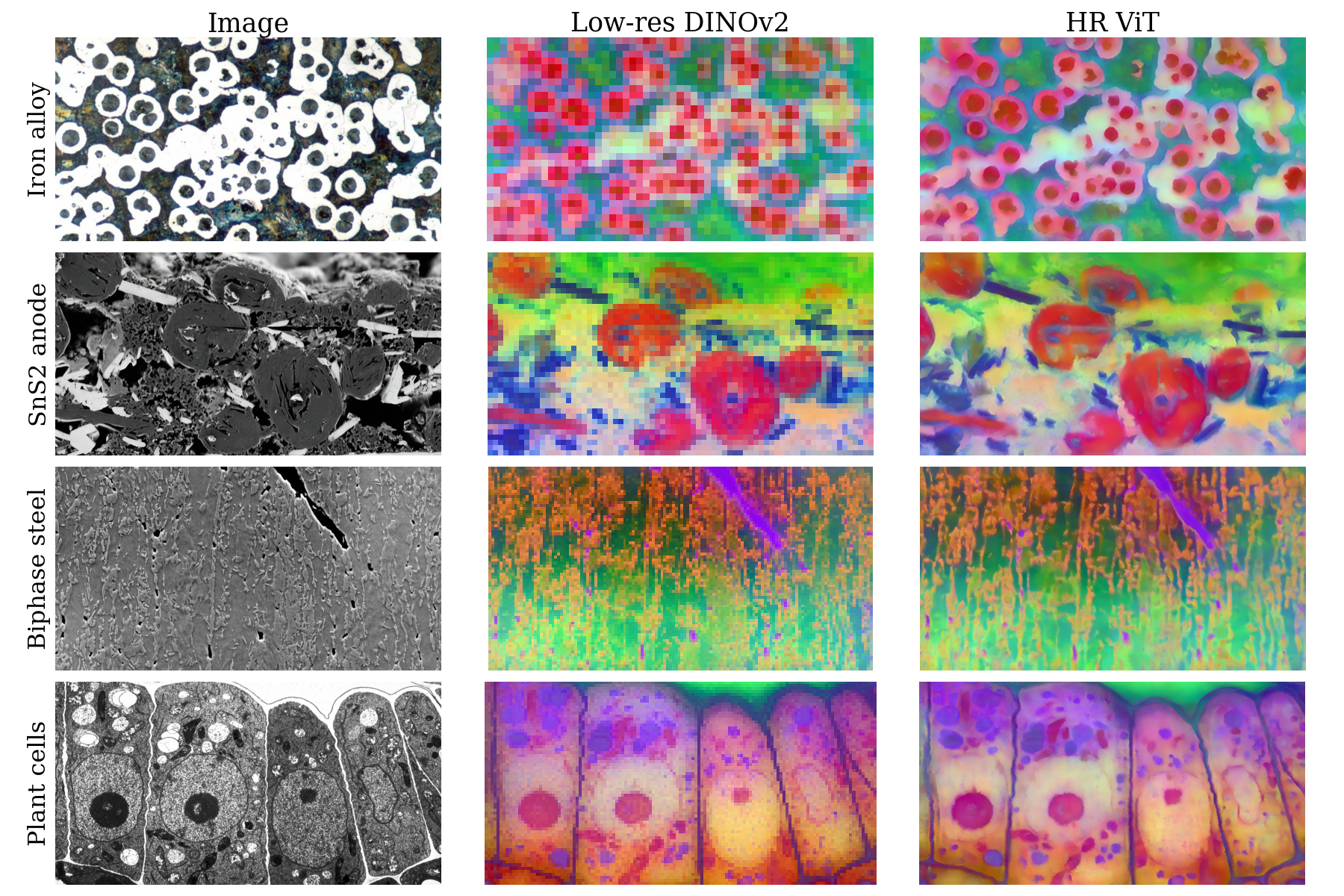}
    \caption{Example low- and high-resolution feature visualisations for a series of micrographs: a 'cast iron alloy with induced spheroidised graphite' \cite{DOITPOMS, DOITPOMS_394}, a 'water-based SnS2/graphite anode'\cite{SNS2_ANODE}, biphase steel \cite{BIPHASE_STEEL} and plant cells \cite{PLANT_SEM}. We see intuitive phase distinctions in the features.} 
    \label{fig:supp:feat_vis}
\end{figure*}

We present more upsampled DINOv2 feature visualisations for micrographs in Figure \ref{fig:supp:feat_vis}, noting that the model tends to display intuitive phase decompositions, even for phases with similar greyscale values. 
For example, the induced graphite in the iron alloy is similar in colour to the iron, but the DINOv2 features identify it as a distinct phase (possibly due to its interiority). 
The same can be seen in the carbon binder vs. graphite particles in the anode, which have the same greyscale values but are structurally distinct. 

\section{Empirical justification of improved performance}
\label{sec:supp:perf_justification}
Our +HR ViT approach appends $k$ channels (usually $k=32$) of upsampled, implicitly PCA'd (see Section \ref{sec:method:feat_preprocess}) ViT features to an existing classical feature-stack. There is the possibility that the improvement in segmentation comes from having additional (redudant) channels to fit over, rather than these channels adding useful information \textit{i.e} they act as regularization. Although it is unlikely for these extra channels to act as a regularization, it is possible if the classifier can only see a subset of the features at any given time, which is the case for the `random forest with feature sampling at splits' formulation of Weka Trainable segmentation.

We test three cases: adding $k=32$ channels of i) random noise, ii) uniform zeros and iii) duplicating the first $k$ channels of the classical feature-stack, and measured segmentation performance on the Ni superalloy benchmark dataset. We report the results in Table \ref{tab:supp:perf_justification}, and find that the each baseline case fails to explain the performance increase of +HR ViT.

\begin{table}[H]
    \centering
    \begin{tabular}{|l|c|}
    \hline
    \textbf{Case} &  \textbf{Class-avg mIoU}  \\
    \hline
    Classical & 0.56 $\pm$ 0.16  \\
    +Random noise   & 0.56 $\pm$ 0.15   \\
    +Uniform  & 0.56 $\pm$ 0.15   \\
    +Duplicated channels  & 0.56 $\pm$ 0.15 \\
    +HR ViT  & \textbf{0.75 $\pm$ 0.11 }\\
    \hline
    \end{tabular}
    \caption{Average mIoU of the three classes relative to the ground truth for Ni superalloy dataset with different feature-stacks concatenated to the classical feature-stack. We see that the uninformative or deleterious additions (uniform zeros, random noise \textit{etc.}) do not improve performance, whereas the HR ViT features do.}
    \label{tab:supp:perf_justification}
\end{table}

Our argument for this improvement is that these features contain additional useful information about the image that maps well to various phase/class distinctions, and allows a classifier to distinguish based on these.
We see evidence of this in the feature map visualisations of Figure \ref{fig:intro_column} and in Section \ref{sec:supp:feature_vis_micrographs}.
For example, imagine a classical setup trying to distinguish cracks inside a particle from narrow gaps between particles - their average colour on various length-scales may be similar, or even identical, and this may make a classifier struggle, and even degrade performance on other pixels. 

If, on the other hand, the HR ViT features contain a notion of `particleness', which is high inside a particle and low around, even in narrow gaps, the classifier can then learn to detect an internal crack using average colour and `particleness', without degrading performance elsewhere. Of course, if there was spurious information in these ViT features the classifier could learn to split on these instead of generalising - we can see an example of such a case in Section \ref{sec:supp:limit_pos_bias}.

\section{Ablation: choice of classifier}
\begin{figure*}
\centering
    \includegraphics[width=0.85\linewidth]{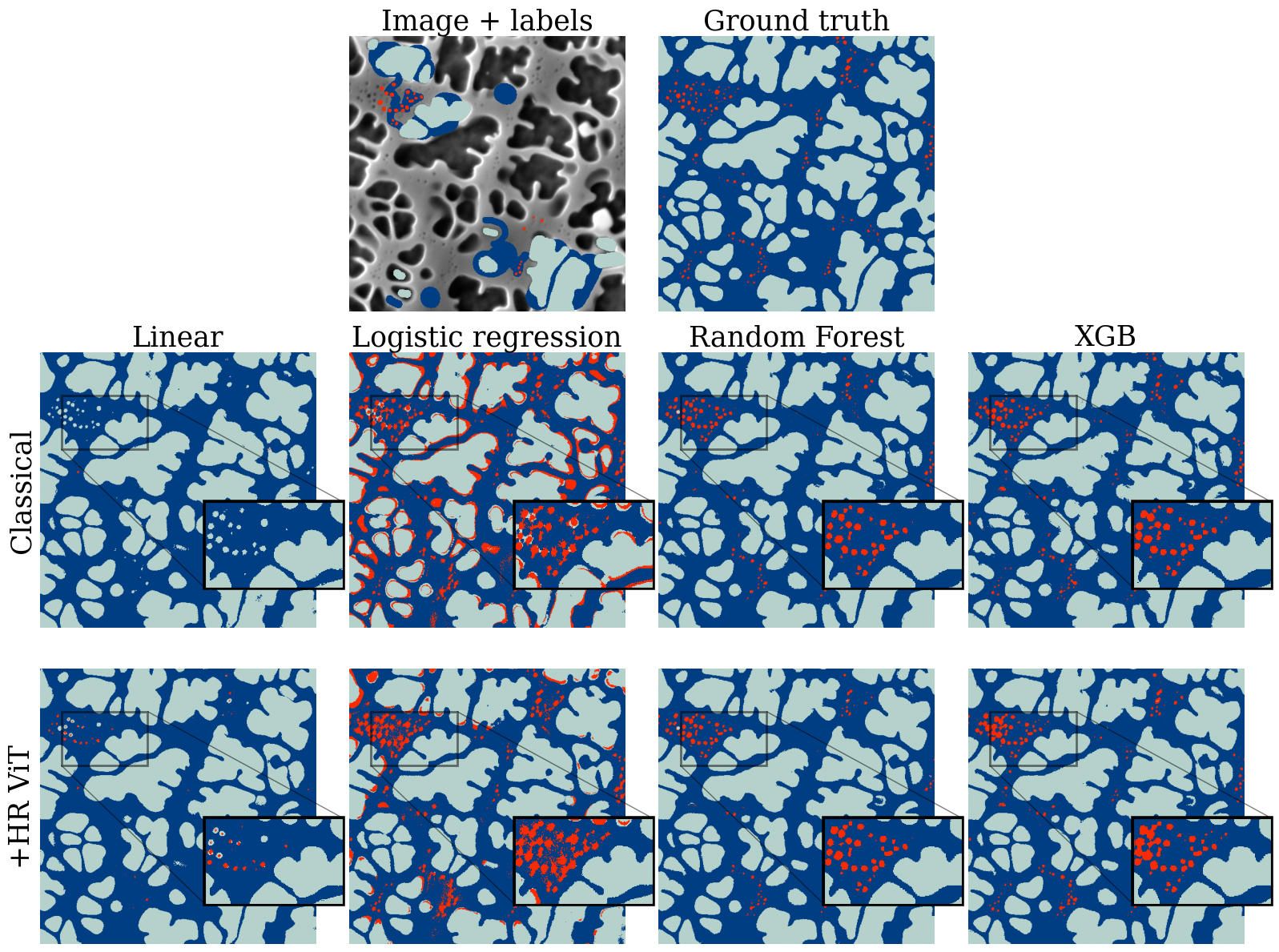}
    \caption{Impact of classifier on interactive segmentation performance with classical and classical + HR ViT features. We see performance increasing with classifier complexity: linear models perform the worst, and tree based models (random forest, XGBoost) perform the best.} 
    \label{fig:supp:classifier_choice}
\end{figure*}

We experiment with changing the classifier used to map from feature-set $\rightarrow$ labels in Figure \ref{fig:supp:classifier_choice}. Tree based methods (XGB and random forest) outperform (generalized) linear models like linear regression and logistic regression. These tree-based methods can model non-linear relationships than linear models cannot (this was the motivation for their use in Trainable Weka Segmentation)\cite{WEKA_PAPER}.

Interestingly, we note that the +HR ViT feature-set can still segment the tertiary precipitates even with a linear model (when the classical features fail) - for this feature-set the non-linearity is implicit in the features, and does not need to be modelled with a non-linear classifier.

\begin{table}[H]
    \centering
    \begin{tabular}{|l|c c|}
    \hline
    \textbf{Classifier} &  \multicolumn{2}{c|}{\textbf{Class-avg mIoU}}  \\
    & Classical & +HR ViT \\
    \hline
    Linear & 0.40 $\pm$ 0.08 & \textbf{0.57 $\pm$ 0.12 }\\
    Logistic   & 0.40 $\pm$ 0.08 & \textbf{0.63 $\pm$ 0.08 }\\
    Random Forest  & 0.55 $\pm$ 0.16 & \textbf{0.68 $\pm$ 0.15 }\\
    XGB   & 0.56 $\pm$ 0.15 & \textbf{0.74 $\pm$ 0.13 }\\
    MLP   & 0.55 $\pm$ 0.11 & \textbf{0.69 $\pm$ 0.12 }\\
    \hline
    \end{tabular}
    \caption{Average mIoU of the three classes relative to the ground truth for Ni superalloy dataset with different classifiers for the different feature-sets.}
    \label{tab:supp:classifier_choice}
\end{table}

We repeat the benchmark experiment of Section \ref{sec:results:is_benchmarks} (4 labelled images, apply to 22, measure mIoU) for the Ni superalloy dataset with a variety of classifiers. XGB and Random Forest perform the best across feature-sets - we use XGB for its faster training speed. We note the relatively strong performance of the linear classifier + HR ViT features setup when compared to the classical feature-set.

\section{Further interactive segmentation examples}

\subsection{On natural images}
\begin{figure*}
\centering
    \includegraphics[width=1\linewidth]{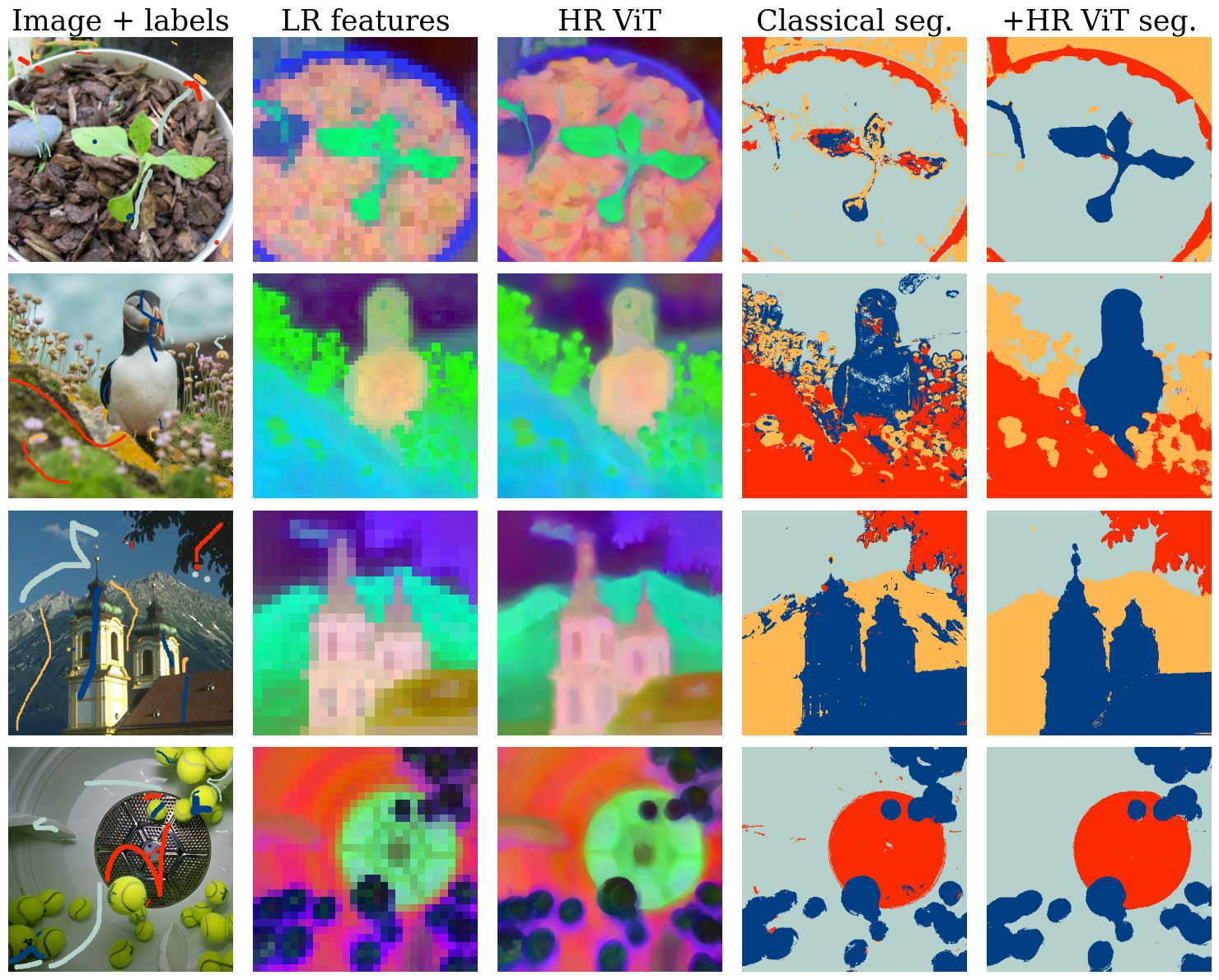}
    \caption{Examples of our approach applied to natural images: again we see good upsamplings of features, which leads to improved segmentations when compared to interactive segmentation using only the classical feature-set. } 
    \label{fig:supp:natural_examples}
\end{figure*}

Figure \ref{fig:supp:natural_examples} shows our approach (feature upsampling + interactive segmentation) applied to a set of natural images (\textit{i.e,} not taken with a microscope). We see good upsamplings of low-res DINOv2 features, this is expected as our upsampler is trained on similar images. We also experiment with performing interactive segmentation on these images, comparing the results when using only the classical feature-set vs the combination of classical and upsampled DINOv2 features. Again we see much improved segmentations as the deep features express strong semantic decompositions.  

\subsection{On micrographs}
\begin{figure*}
\centering
    \includegraphics[width=1\linewidth]{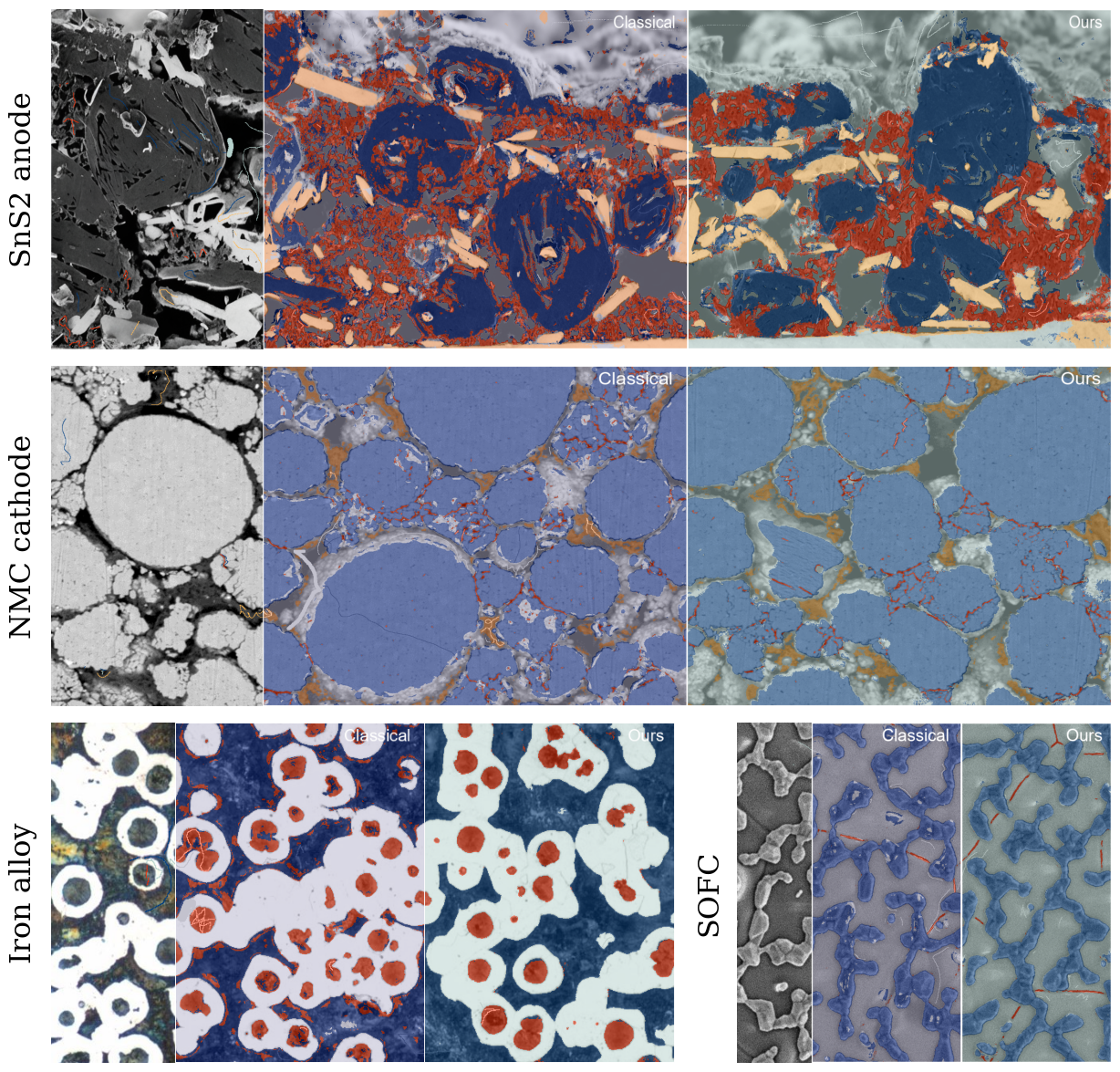}
    \caption{Further interactive segmentation examples on micrographs - we compare segmentations produced using classical features vs classical + upsampled ViT features (`ours`). This includes a 'water-based SnS$_2$/graphite anode'\cite{SNS2_ANODE}, an NMC cathode\cite{KINTSUGI}, a 'cast iron alloy with induced spheroidised graphite' \cite{DOITPOMS, DOITPOMS_394}.  } 
    \label{fig:supp:more_micrographs}
\end{figure*}

We show further examples of our approach improving micrograph segmentation in Figure \ref{fig:supp:more_micrographs}. The active material (blue) and carbon binder (red) distinction is greatly improved for a tin-sulfide anode \cite{SNS2_ANODE}, as well as improved out-of-plane / in-pore differentiation. This ability to distinguish between in-plane (flat) and out-of-plane (textured) material is important for resolving the `pore-back' problem in imaging of porous media, we see a further example of this when applied to an NMC cathode SEM\cite{KINTSUGI} in the second row. The classical features struggle to consistently differentiate between in-plane active material (blue) and out-of-plane material (grey), especially when compared to the carbon binder (yellow). 

This improvement continues for a relfected light micrograph of cast iron alloy \cite{DOITPOMS_394}, which contains spherodised graphite (red). The classical features struggle to separate the graphite from the iron matrix (blue), due to a similar (average) colour - the DINOv2 features are able to capture this distinction (see Section \ref{sec:supp:feature_vis_micrographs} for an explanation).  

\section{Predictions of fully trained classifiers}

\begin{figure*}
\centering
    \includegraphics[width=0.8\linewidth]{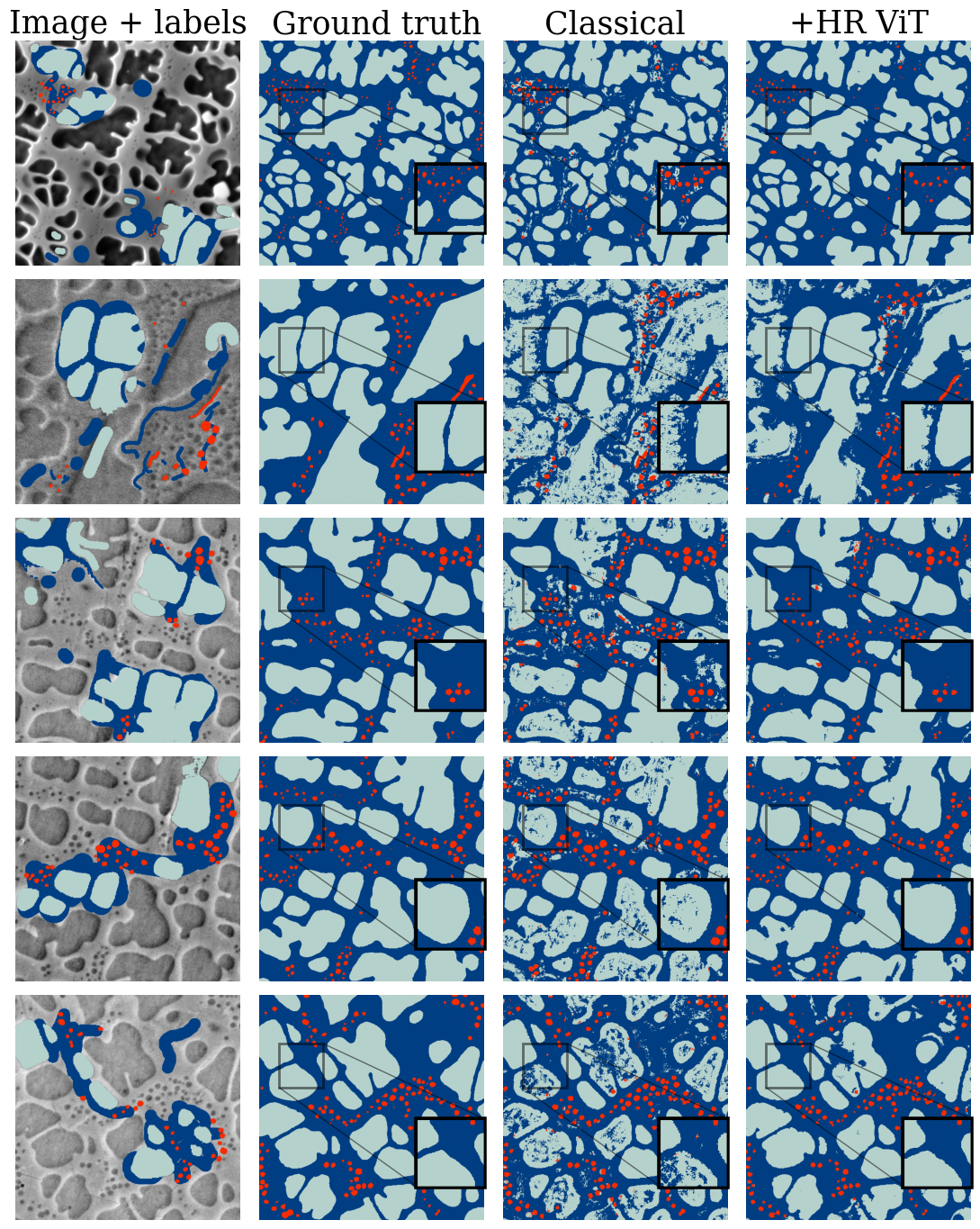}
    \caption{Example segmentation results for `fully-trained' classifiers \textit{i.e,} trained on labels across the whole dataset of 22 Ni superalloys. The same labels and classifier type were used for each feature-set. We see that adding the HR ViT features greatly improves the accuracy of the resulting segmentations. }
    \label{fig:supp:max_label_preds}
\end{figure*}

Figure \ref{fig:supp:max_label_preds} shows predictions of classifiers trained on labels across all 22 images in the dataset with either the Classical or +HR ViT dataset. Each image is shown with its labels (similar labels were applied to other unshown images) and the associated ground truth. The same labels were used for both feature-sets. We see that the +HR ViT features produce much better segmentations, with fewer mispredictions across each classes. Both approaches get the gross structure correct, but the noisiness of the Classical feature-set's predictions make them unsuitable for counting precipitate size, or using in downstream simulations.

\section{Limitations: positional bias}
\label{sec:supp:limit_pos_bias}
ViTs have a positional encoding (see Section \ref{sec:theory:vits}) added to each patch token which represents where that patch has come from in the image. This allows ViTs to express features over objects that span multiple patches, but also leads to this information being persisted into the resulting feature-stacks. This is undesirable, but can compensated for by labelling homogenously or by averaging over various transformations (flips \textit{etc.}) \cite{HR-DV2}. Work has been done on correcting similar artefacts in vision transformers \cite{DENOISING_VISION_TRANSFORMERS}.

\begin{figure*}
\centering
    \includegraphics[width=\linewidth]{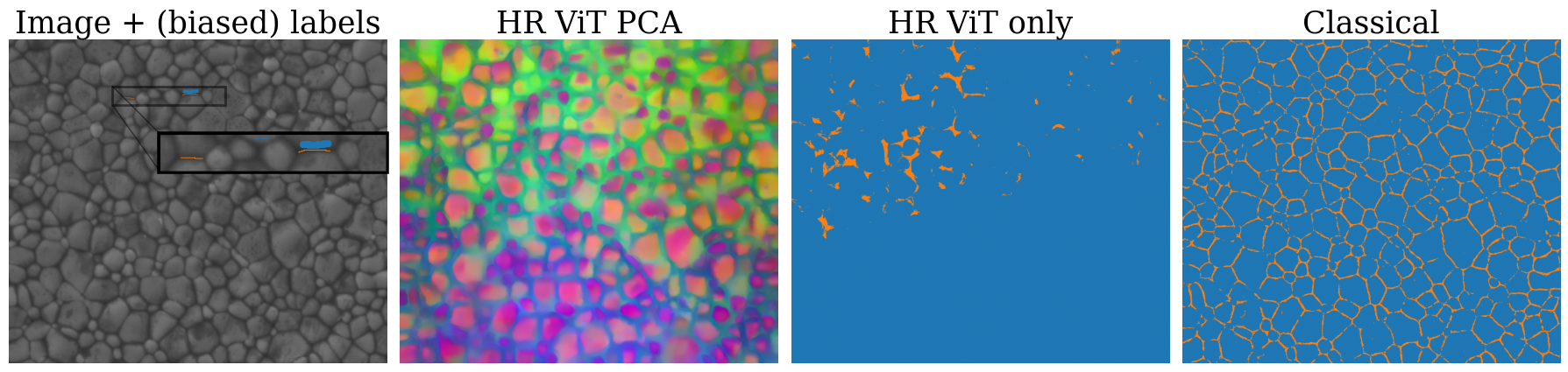}
    \caption{The positional bias problem of HR ViT features. We see a clear top-to-bottom gradient in the PCA of the HR ViT feature-stack, which is reflected when we try to train a classifier on positionally biased labels. The classical features, being homogenous, do not suffer from this problem. This can be compnesated with spatially unbiased labels, or by homognenising the deep features by averaging over flip transforms. } 
    \label{fig:supp:pos_bias}
\end{figure*}

We show the issues this causes for interactive segmentation in Figure \ref{fig:supp:pos_bias}. There is a clear gradient in the PCA visualisation of the HR ViT feature-stack, which results in a poor segmentation when a classifier is trained on only the HR ViT features (note: this is not the same case as +HR ViT, which refers to concatenating the ViT features to the classical ones). The homogenous classical features do not struggle, even with biased labels. Further discussion, alongside an example of using flip transforms to ameliorate this problem, can be found in ref. \cite{HR-DV2}.

